\documentclass{article}

\usepackage[preprint]{neurips_2024}

\usepackage[utf8]{inputenc} %
\usepackage[T1]{fontenc}    %
\usepackage{hyperref}       %
\usepackage{url}            %
\usepackage{booktabs}       %
\usepackage{amsfonts}       %
\usepackage{nicefrac}       %
\usepackage{microtype}      %
\usepackage{xcolor}         %
\usepackage{graphicx}
\usepackage{wrapfig}
\usepackage{amsmath}
\usepackage{amssymb}
\usepackage{mathtools}
\usepackage{amsthm}
\usepackage{bm}
\usepackage{algorithm}
\usepackage{algpseudocode}
\setcitestyle{round}
\hypersetup{
   colorlinks=true,
   linkcolor=[RGB]{30, 30, 180},
   citecolor=[RGB]{30, 30, 180},
   urlcolor=magenta,
   pdfborder=0 0 0,
   pdftitle={},
   pdfsubject={}, 
   pdfkeywords={},
   pdfauthor={},%
   pdfstartview=FitH
}
\usepackage[capitalize,noabbrev]{cleveref}
\usepackage{xspace}
\usepackage{array}

\newcommand{\OurMethod}{ReCap\xspace}
\newcommand{\DAL}{ReCap+DAL\xspace}

\newcommand\Bc{\bm{c}}

\newcommand\Be{\bm{e}}
\newcommand\Bf{\bm{f}}

\newcommand\Bp{\bm{p}}
\newcommand\Bq{\bm{q}}
\newcommand\Br{\bm{r}}
\newcommand\Bs{\bm{s}}

\newcommand\Bx{\bm{x}}

\newcommand\BE{\bm{E}}
\newcommand\BF{\bm{F}}

\newcommand\BI{\bm{I}}

\newcommand\BU{\bm{U}}
\newcommand\BV{\bm{V}}
\newcommand\BW{\bm{W}}

\newcommand\BSi{\bm{\Sigma}}

 \newcommand{\dN}{\mathbb{N}}
  
 \newcommand{\dR}{\mathbb{R}}

\newcommand{\cC}{\mathcal{C}} \newcommand{\cD}{\mathcal{D}}

\newcommand{\cK}{\mathcal{K}} 
 
\newcommand{\cO}{\mathcal{O}} 
 \newcommand{\cR}{\mathcal{R}}
\newcommand{\cS}{\mathcal{S}} \newcommand{\cT}{\mathcal{T}}
 
 \newcommand{\cX}{\mathcal{X}}

\makeatletter
\newcommand{\dlmf}[1]{%
\citep[%
  \def\nextitem{\def\nextitem{, }}%
  \@for \el:=#1\do{\nextitem\href{http://dlmf.nist.gov/\el}{(\el)}}%
]{olver_nist_2010}%
}
\makeatother

\title{Linear Alignment of Vision-language Models for Image Captioning}

 \author{
  \textbf{Fabian Paischer}$~^{1}$,
  \textbf{Markus Hofmarcher}$~^{2}$,
  \textbf{Sepp Hochreiter}$~^{1}$,
  \textbf{Thomas Adler}$~^{1}$\\
  $~^{1}$~ELLIS Unit Linz and LIT AI Lab, Institute for Machine Learning,\\
  $~^{2}$~JKU LIT SAL eSPML Lab, Institute for Machine Learning,\\
  Johannes Kepler University, Linz, Austria\\
  \texttt{paischer@ml.jku.at}
}

\begin{document}

\maketitle

\begin{abstract} 
    Recently, vision-language models like CLIP have advanced the state of the art in a variety of multi-modal tasks including image captioning and caption evaluation. 
    Many approaches leverage CLIP for cross-modal retrieval to condition pre-trained language models on visual input.
    However, CLIP generally suffers from a mis-alignment of image and text modalities in the joint embedding space.
    We investigate efficient methods to linearly re-align the joint embedding space for the downstream task of image captioning.
    This leads to an efficient training protocol that merely requires computing a closed-form solution for a linear mapping in the joint CLIP space.
    Consequently, we propose a lightweight captioning method called \OurMethod, which can be trained up to 1000 times faster than existing lightweight methods. 
    Moreover, we propose two new learning-based image-captioning metrics built on CLIP score along with our proposed alignment.
    We evaluate \OurMethod on MS-COCO, Flickr30k, VizWiz and MSRVTT. 
    On the former two, \OurMethod performs comparably to state-of-the-art lightweight methods using rule-based metrics while outperforming them on most of the CLIP-based metrics. 
    On the latter two benchmarks, \OurMethod consistently outperforms competitors across all metrics and exhibits strong transfer capabilities and resilience to noise.
    Finally, we demonstrate that our proposed metrics correlate stronger with human judgement than existing metrics on the Flickr8k-Expert, Flickr8k-Crowdflower, and THumB datasets.
\end{abstract}

\section{Introduction}
Vision-language models (VLMs) are usually trained to align images and texts in a joint bi-modal embedding space.
As one of the most prominent VLMs, CLIP \citep{radford_learning_2021} has been pre-trained on a large-scale web dataset consisting of image-text pairs and advanced the state of the art across a variety of vision-language tasks.
These tasks include, but are not limited to image-text retrieval \citep{ramos_smallcap_2023}, image captioning \citep{mokady_clipcap_2021}, few-shot classification \citep{ouali_black_2023}, and caption evaluation \citep{hessel_clipscore_2021}.
One of the most important downstream tasks is image captioning.
It requires machines to generate informative descriptions of images which can be useful in various applications, such as content-based image search, or accessibility for visually impaired individuals \citep{gurari_captioning_2020}.

\begin{figure*}[h]
  \begin{center}
    \includegraphics[width=\textwidth]{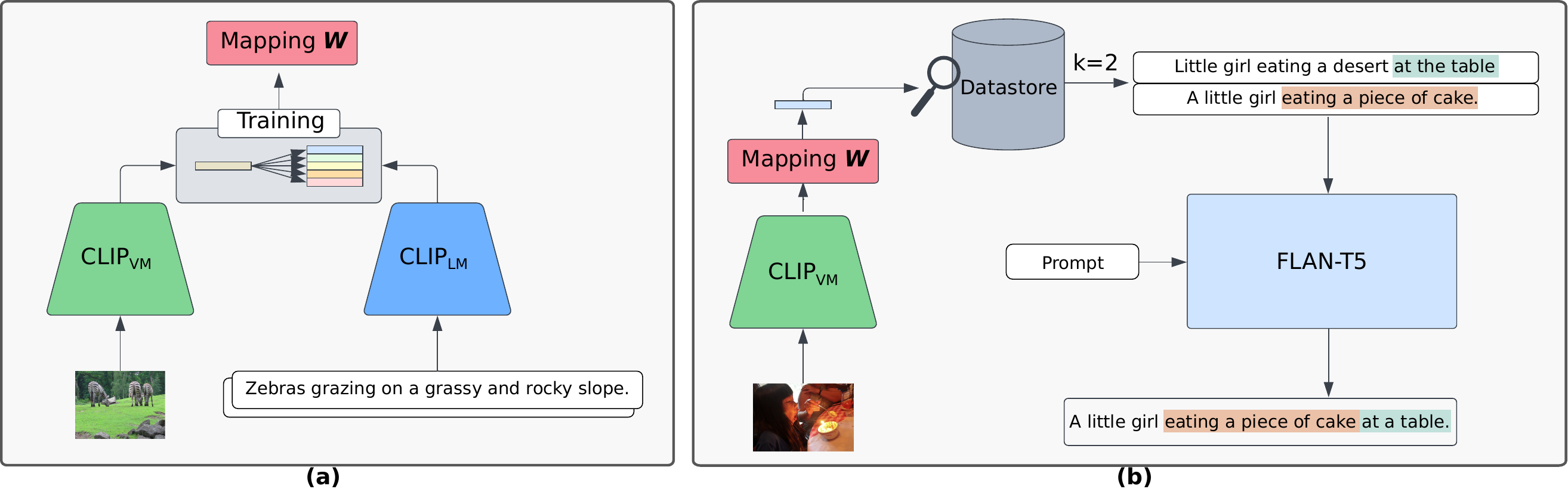}
  \end{center}
  \caption{\textbf{(a)} We train a linear mapping $\BW$ to align the image and text embeddings of CLIP toward a dataset. \textbf{(b)} On inference, we employ the mapping to retrieve captions from a datastore that are similar to the input image and provide these along with a prompt to a FLAN-T5 model to generate a new caption.}
  \label{fig:methodology}
\end{figure*}

CLIP suffers from a mis-alignment between image and text modalities in its joint embedding space \citep{liang_mind_2022}. 
Adapting CLIP to a downstream task is generally costly in terms of both computational resources and data collection. 
Therefore, we explore efficient ways to re-align image and text embeddings of CLIP-style models to leverage them for retrieval augmentation for image captioning.
This use case of CLIP is based on cross-modal retrieval via cosine similarity.
The globally optimal linear solution to a constrained least-squares problem is equivalent to maximizing the cosine similarity under the same constraint \citep{artetxe_learning_2016}.
Leveraging this insight, we maximize the cosine similarity of image-text correspondences from the downstream dataset with respect to a constrained linear mapping. 
As this problem has a closed-form solution, we are able to align CLIP to the downstream data without backpropagation. 
This makes our proposed method extremely fast and versatile as training takes only seconds and can be conducted on CPU. 

We propose a fast and easily deployable method for adapting CLIP to a target domain. 
Given a set of image-text pairs representing a downstream task, we embed them in the joint embedding space of CLIP.
Then we re-align them by computing a linear mapping via a constrained least-squares solution (cf.~\cref{fig:methodology}, a).
The linear mapping introduces only 0.0016\% of trainable parameters compared to the original CLIP model.
We demonstrate that this technique can be readily incorporated into an image captioning pipeline via retrieval augmentation (cf.~\cref{fig:methodology}, b). 
Given a new image, we embed it in the CLIP embedding space and apply our mapping before retrieving similar captions via cosine similarity.
These captions are then formatted to a prompt which is provided to a LM to generate a new caption for the image.
We call the resulting method \textbf{Re}trieval-augmented \textbf{Cap}tioner (\OurMethod).
Further, since established image-captioning evaluation metrics mostly rely on rule-based matching to reference captions \citep{papineni_bleu_2002,vedantam_cider_2015}, we  propose two new learning-based image-captioning metrics that use our linear alignment to adapt CLIP-based metrics \citep{hessel_clipscore_2021} toward a downstream dataset.

We evaluate \OurMethod on the MS-COCO \citep{lin_microsoft_2014}, Flickr30k \citep{young_from_2014}, VizWiz \citep{gurari_captioning_2020}, and MSRVTT \citep{xu_msrvtt_2016} datasets. 
By means of rule-based metrics, \OurMethod achieves performance competitive to lightweight baselines that require over 1000 times more training effort on MS-COCO and Flickr30k, while outperforming other lightweight competitors on VizWiz and MSRVTT. 
By means of CLIP-based metrics including those proposed in this work, \OurMethod mostly performs on-par or better than competitors on all four datasets. 
Additionally, we present evidence that \OurMethod can leverage out-of-distribution data for retrieval more effectively than other lightweight retrieval augmented methods.
Further, we evaluate the correlation of our proposed metrics with human judgement on three datasets, Flickr8k-Expert and Flickr8k-Crowdflower \citep{hodosh_framing_2013}, and THumB \citep{kasai_transparent_2022}.
Our metrics improve over the CLIP-based metrics that rely on cosine similarity \citep{hessel_clipscore_2021} on average across all datasets.

\section{Methods}
\label{sec:method}

We propose a linear alignment method for CLIP that optimizes cosine similarity between image-text pairs coming from a downstream dataset.
The linear alignment constitutes a closed-form linear mapping.
Therefore, it is very efficient to compute and easy to implement while only adding a relatively small set of trainable parameters.
We elaborate on our linear alignment technique in more detail in \cref{sec:close_gap}.
In \cref{sec:method_captioning} we introduce a lightweight image-captioning pipeline based on our linear alignment without any further training.
Finally, \cref{sec:captioning_metric} introduces two new image-captioning metrics, aCLIP-S, a reference-free metric, and RefaCLIP-S, a reference-based metric, both of which are based on the CLIP score \citep{hessel_clipscore_2021} in combination with our proposed linear alignment.

\subsection{Linear Alignment of CLIP}
\label{sec:close_gap}

Since our downstream use of CLIP involves retrieval via cosine similarity, we want to maximize the cosine similarity between image and text embeddings of a downstream dataset. 
To this end, we assume access to a dataset  $\cD = \{(\Bx_i, \Bc_i)\}$ that provides image-text pairs.
First, we embed the images of the training split $\cD_{\text{Train}} \subset \cD$ using a CLIP vision encoder $\phi: \cX \to \dR^d$, where $\cX$ is the pixel space and $d$ denotes the dimension of the joint CLIP embedding space.
This results in an image embedding matrix $\BF_{\cD_{\text{Train}}} = (\Bf_1, \dots, \Bf_n)^\top  \in \dR^{n \times d}$, where $\Bf_i = \phi(\Bx_i)$ for $i \in \{1, \dots, n\}$ and $n = |\cD_{\text{Train}}|$.
Similarly, we embed the corresponding captions via the CLIP text encoder $\psi: \cT \to \dR^d$, where $\cT$ is the space of tokenized strings, yielding a caption embedding matrix $\BE_{\cD_{\text{Train}}} = (\Be_1, \dots, \Be_n)^\top \in \dR^{n \times d}$. 
If, like in the case of MS-COCO, we are presented with multiple captions per image, then we assume the same image just appears multiple times in $\cD$ matched up with its corresponding captions. 
This results in a one-to-one correspondence between inputs and labels. 

We employ a linear mapping $\BW \in \dR^{d \times d}$ to re-align CLIP according to $\cD_{\text{Train}}$. 
We aim to find a mapping $\BW$ that projects an image embedding to the text embedding space such that its closest neighbor in terms of cosine similarity is its ground-truth caption. 
Yet, a closed-form solution for $\BW$ to maximize the cosine similarity is unknown. 
By constraining $\BW$ to be an orthogonal matrix, however, we obtain equivalence to the least-squares objective, that is
\begin{equation}\label{eq:procrustes}
    \BW^{\ast}=
    \operatorname*{arg\,max}_{\BW \text{ s.t. } \BW^\top \BW = \BI} \sum_i \operatorname{cossim}(\Be_i, \BW \Bf_i)
    =\operatorname*{arg\,min}_{\BW \text{ s.t. } \BW^\top \BW = \BI} \sum_i \| \Be_i - \BW \Bf_i \|_2^2 = \BV \BU^\top,
\end{equation}
where $\BV$ and $\BU$ are the orthogonal matrices of the singular value decomposition of $\BE_{\cD_{\text{Train}}}^\top \BF_{\cD_{\text{Train}}} = \BU \BSi \BV^\top$ and $\operatorname{cossim}(\cdot, \cdot)$ is the usual cosine similarity for vectors.
This fact was shown by \citet{artetxe_learning_2016} and we also provide a proof in \cref{app:motivation} for convenience.
The solution to the constrained optimization problem in \cref{eq:procrustes} is well known as \textit{orthogonal procrustes} in the literature \citep{schonemann_generalized_1966}.
Since the size of $\BW$ depends on $d$, the dimension of the embedding space, different CLIP encoders result in different numbers of parameters introduced by $\BW$.

\subsection{Retrieval-augmented Image Captioning (ReCap)}
\label{sec:method_captioning}

Our linear mapping $\BW$ can be leveraged for task-specific alignment and gives rise to our novel lightweight image captioning method \OurMethod.
The key idea is that we can represent a given image in the language space as a set of captions that describe similar images.
To this end, we utilize a datastore of embedded captions from which we can retrieve. 
In turn, we can condition a pre-trained language model (LM) on this set of retrieved captions to create a new caption for the input image.

We utilize $\BW$ for retrieval augmentation, where the retrieval datastore $\cC$ contains captions of the training set $\cD_{\text{Train}}$.
Then we project a given image to the caption embedding space and retrieve its nearest neighbors.
Given an image $\Bx \in \cX$, we compute an embedding $\phi(\Bx)$ and select the set $\cK$ of top-$k$ captions by
\begin{equation} \label{eq:topk}
    \cK = \operatorname*{arg\,max}^k_{\Bc \in \cC} \operatorname{cossim}(\psi(\Bc), \BW \phi(\Bx)),
\end{equation}
where $\arg \max^k$ denotes an extension of the $\arg \max$ operator returning the arguments of the $k$ largest elements of a set. 
This way, we obtain a set of captions that provide a textual description of the image $\Bx$.
We feed the retrieved captions $\cK$ to a generative LM as context along with a prompt to generate a new caption for the image $\Bx$ (cf.~\cref{fig:methodology}, b).
We use nucleus sampling \citep{holtzman_curious_2020} to obtain a set $\cS$ of $l$ candidate captions for the image $\Bx$ and select the candidate which yields the highest cosine similarity by
\begin{equation}\label{eq:filtering}
    \operatorname*{arg\,max}_{\Bs \in \cS} \operatorname{cossim}(\psi(\Bs), \BW \Bf).
\end{equation}
The only trainable parameters of \OurMethod are $\BW$ which only requires computing a closed-form solution on CPU.
Specifically, computing $\BW$ requires $\cO(d^3)$ steps. 
The function \textsc{ReCap} in \cref{alg:self_improvement} shows pseudocode for our lightweight image-captioning method.

\subsection{Image Caption Evaluation Metric}
\label{sec:captioning_metric}

Given an image $\Bx$ and a candidate caption $\Bc$ we define the aligned CLIP score as
\begin{equation}
    \operatorname{aCLIP-S}(\Bc, \Bx) = \max \{\operatorname{cossim}(\psi(\Bc), \BW \phi(\Bx)), 0\}.
\end{equation}
Notably, aCLIP-S is reference-free, meaning it can be applied to any candidate without access to ground-truth human annotations, i.e. reference captions.
In case a set $\cR = \{\Br_1, \Br_2, \dots\}$ of reference captions is available, we can incorporate those into our score, which results in a reference-based metric
\begin{equation}
    \operatorname{RefaCLIP-S}(\Bc, \cR, \Bx) = 
    \operatorname{H}\{\operatorname{aCLIP-S}(\Bc, \Bx), \max\{ \max_{\Br \in \cR} \operatorname{cossim}(\psi(\Bc), \psi(\Br)), 0\} \},
\end{equation}
where $\operatorname{H}\{\cdot\}$ denotes the harmonic mean of a set.
Since our new metrics use data to align CLIP to the downstream task, we categorize them as learning-based \citep{cui_learning_2018}.

\section{Experiments}
\label{sec:experiments}

First, we show results for \OurMethod on the common captioning benchmarks MS-COCO \citep{lin_microsoft_2014} and Flickr30k \citep{young_from_2014} in \cref{sec:exp_captioning}.
To investigate how \OurMethod copes with noisy data and video captions, we additionally show results for the VizWiz \citep{gurari_captioning_2020} and MSRVTT \citep{xu_msrvtt_2016} datasets.
Moreover, we investigate the transfer capabilities of (i) our linear mapping alone and (ii) of mapping and datastore combined across different domains.
In \cref{sec:experiments_metric} we evaluate our proposed image captioning metrics on the Flickr8k-Expert and Flickr8K-Crowdflower \citep{hodosh_framing_2013}, and  the THumB dataset \citep{kasai_transparent_2022}.
Finally,  we evaluate different linear alignment methods on cross-modal retrieval on MS-COCO and Flickr30k benchmarks and contrast their performance to their unaligned counterparts in \cref{sec:cross_modal_retrieval}.

\subsection{\OurMethod}
\label{sec:exp_captioning}
We leverage retrieval augmentation to enable caption generation via a generative LM.
This results in an extremely efficient training protocol which merely requires computation of the linear mapping to align the pre-trained CLIP.

\paragraph{Implementation Details}
During downstream evaluation of our linear alignment we rely on cosine similarity for retrieval of texts related to an image.
Therefore, we evaluate all CLIP vision encoders on cross-modal retrieval tasks in \cref{sec:additional_results} to find a suitable encoder for \OurMethod.
Based on our findings, we choose RN50$\times$64 \citep{he_deep_2016} as our retrieval model.\footnote{We take the RN50$\times$64 model from the official repository at \url{https://github.com/openai/CLIP}.}
After embedding images and captions we normalize and center them as suggested by \citet{artetxe_learning_2016}.
To compute our mapping, we use orthogonal procrustes by default as described by \cref{eq:procrustes}. 
In certain settings, we use an unconstrained version, i.e., ordinary least squares.
We elaborate in \cref{sec:additional_results} which version we use for the different experiments.

To find the best setting for image captioning, we search over different LMs, decoding strategies, and prompt orderings.
We only considered generative LMs that are publicly available on the huggingface hub \citep{wolf_transformers_2020}.
Moreover, we search over multiple values of retrieved captions ($k$).
We always search hyperparameters on the validation split of the respective dataset.
For more details about hyperparameters, see \cref{app:hyperparams}.
We use \texttt{faiss} \citep{johnson_billion_2019} to manage our datastore since it enables efficient storage and retrieval of vectors.
Our final setting uses a FLAN-T5-Large \citep{chung_scaling_2022} with nucleus sampling.
To generate captions with FLAN-T5, we explore different prompting strategies and found the strategy proposed in \citet{ramos_smallcap_2023} to work best.
Specifically, we use the prompt template ``\textit{Similar images show:} $<\text{caption}_1>,\ldots,<\text{caption}_k>$ \textit{This image shows:}''.

\begin{table*}[t]
\caption{Comparison of different lightweight methods on the MS-COCO, Flickr30k, VizWiz, and MSRVTT test sets. We report round mean and standard error and mark results we computed ourselves with an asterisk. We omit error bars when they are not available.}
\label{tab:caption_all}
\begin{center}
\begin{small}
\begin{sc}
\resizebox{\textwidth}{!}{
    \begin{tabular}{l | c c c c c c c c }
    \toprule
    & \multicolumn{8}{c}{MS-COCO} \\
    Method & CIDEr-D & SPICE & CLIP-S & RefCLIP-S & CLIP+DN & CLIP+DN-Ref & aCLIP-S & RefaCLIP-S \\
    \midrule
    ClipCap*  & 103.8 $\pm$ 1.0 & 19.9 $\pm$ 0.1 & 74.6 $\pm$ 0.1 & 79.9 $\pm$ 0.1  & 18.6 & 40.2 $\pm$ 0.1 & \textbf{46.1 $\pm$ 0.1} & 57.5 $\pm$ 0.1 \\
    $\text{I-Tuning}_{\text{Base}}$ & 116.7 & \textbf{21.8} & n/a & n/a & n/a & n/a & n/a & n/a \\
    Prefix-Diffusion & 106.3 & 19.4 & 63.4 & 70.9 & n/a & n/a & n/a & n/a \\
    $\text{SmallCap}_{\text{d=4,Base}}$* & \textbf{117.6 $\pm$ 1.0} & 21.1 $\pm$ 0.1 & \textbf{75.1 $\pm$ 0.1} & \textbf{80.5 $\pm$ 0.}1 & \textbf{18.8 $\pm$ 0.1} & \textbf{40.6 $\pm$ 0.1} & \textbf{46.1 $\pm$ 0.1} & 57.7 $\pm$ 0.1 \\
    $\text{ReCap}$ (Ours)* & 108.3 $\pm$ 1.0 &  21.2 $\pm$ 0.1 & 74.3 $\pm$ 0.1  & \textbf{80.4 $\pm$ 0.1} & \textbf{18.6 $\pm$ 0.1} & \textbf{40.6 $\pm$ 0.1} & \textbf{46.1 $\pm$ 0.1} & \textbf{58.0 $\pm$ 0.1}  \\ %
    \midrule
    & \multicolumn{8}{c}{Flickr30k} \\
    \midrule
    ClipCap*  & 57.0 $\pm$ 1.8 & 15.8 $\pm$ 0.3 & 73.8 $\pm$ 0.3 & 75.9 $\pm$ 0.3 & 16.5 $\pm$ 0.1 & 36.3 $\pm$ 0.2 & 44.1 $\pm$ 0.2 & 53.0 $\pm$ 0.2 \\
    $\text{I-Tuning}_{\text{Base}}$ & 61.5 & 16.9 & n/a & n/a & n/a & n/a & n/a & n/a \\
    Prefix-Diffusion & 53.8 & 14.2 & 61.6 & 66.3 & n/a & n/a & n/a & n/a \\
    $\text{SmallCap}_{\text{d=4,Base}}$*  &  \textbf{69.6 $\pm$ 2.1} & \textbf{17.1 $\pm$ 0.3} & \textbf{75.8 $\pm$ 0.3} & 78.2 $\pm$ 0.2 & 17.3 $\pm$ 0.1 & 37.7 $\pm$ 0.2 & \textbf{44.1 $\pm$ 0.2} & \textbf{55.0 $\pm$ 0.2} \\ 
    $\text{ReCap}$ (Ours)* & \textbf{68.8 $\pm$ 2.0} & \textbf{17.5 $\pm$ 0.3} & 
    \textbf{76.1 $\pm$ 0.2} & \textbf{79.4 $\pm$ 0.2} & \textbf{17.9 $\pm$ 0.1} & \textbf{38.8 $\pm$ 0.1} & \textbf{44.1 $\pm$ 0.2} & \textbf{55.0 $\pm$ 0.2} \\ %
    \midrule
    & \multicolumn{8}{c}{VizWiz} \\
    \midrule
    ClipCap* & 48.1 $\pm$ 0.0 & 13.4 $\pm$ 0.0 & 69.7 $\pm$ 0.1 & n/a & 13.7 $\pm$ 0.0 & n/a & 20.1 $\pm$ 0.1  & n/a \\
    $\text{SmallCap}_{\text{d=4,Base}}$* & 51.9 $\pm$ 0.0 & 13.4 $\pm$ 0.0 & \textbf{75.0 $\pm$ 0.1} & n/a & \textbf{15.6 $\pm$ 0.1} & n/a & 21.6 $\pm$ 0.1 & n/a \\ 
    $\text{ReCap}$ (Ours)* & \textbf{62.3 $\pm$ 0.0} & \textbf{16.7 $\pm$ 0.0} & 73.5 $\pm$ 0.1 & n/a  & \textbf{15.5 $\pm$ 0.1} & n/a & \textbf{26.6 $\pm$ 0.1} & n/a \\
    \midrule
    & \multicolumn{8}{c}{MSRVTT} \\
    \midrule
    ClipCap* & 2.0 $\pm$ 0.0 & 10.4 $\pm$ 0.0 & 64.2 $\pm$ 0.0 & 68.7 $\pm$ 0.0 & 10.9 $\pm$ 0.0 & 29.6 $\pm$ 0.0 & 23.8 $\pm$ 0.0 & 31.5 $\pm$ 0.0 \\
    $\text{SmallCap}_{\text{d=4,Base}}$* & 31.6 $\pm$ 0.2 & 11.1 $\pm$ 0.0 & 57.1 $\pm$ 0.0 & 65.0 $\pm$ 0.0 & 7.5 $\pm$ 0.0 & 26.7 $\pm$ 0.0 & 22.1 $\pm$ 0.0 & 30.2 $\pm$ 0.0 \\ 
    $\text{ReCap}$ (Ours)* & \textbf{38.8 $\pm$ 0.2} & \textbf{14.4 $\pm$ 0.0}  & \textbf{67.6 $\pm$ 0.0} & \textbf{71.1 $\pm$ 0.0} & \textbf{12.8 $\pm$ 0.0} & \textbf{31.8 $\pm$ 0.0} & \textbf{25.6 $\pm$ 0.0} & \textbf{35.1 $\pm$ 0.0} \\
    \bottomrule
\end{tabular}
}
\end{sc}
\end{small}
\end{center}
\vskip -0.1in
\end{table*}

\paragraph{Datasets}
We split the MS-COCO and Flickr30k benchmarks according to \citet{karpathy_deep_2017} into train, validation, and test splits.
For MSRVTT and VizWiz we split according to the official splits \citep{gurari_captioning_2020,xu_msrvtt_2016}.
Since VizWiz contains a substantial amount of noise, we filter out all captions for images that suffer from severe quality issues or were rejected by annotators and evaluate the generated test captions on the official evaluation server.\footnote{\url{https://eval.ai/web/challenges/challenge-page/739/overview}}
For MSRVTT, we employ the same pre-processing pipeline as \citet{ramos_smallcap_2023} and extract four frames from each video and pair them with the ground truth captions.
This results in many-to-many correspondences.

\paragraph{Baselines}
We consider existing methods as lightweight if their trainable parameter count is below 50\,M.
For MS-COCO and Flickr30k, we compare \OurMethod to ClipCap \citep{mokady_clipcap_2021}, I-Tuning \citep{lup_ituning_2023}, SmallCap \citep{ramos_smallcap_2023}, and Prefix-Diffusion \citep{liu_prefix_2023}.
For MSRVTT and VizWiz, we compare \OurMethod to SmallCap, since it is the only existing lightweight method that report results on these datasets.
We report implementation details about the baselines in \cref{sec:app_impl_details}.

\begin{wrapfigure}{r}{0.5\textwidth}
    \begin{center}
    \includegraphics[width=.5\textwidth]{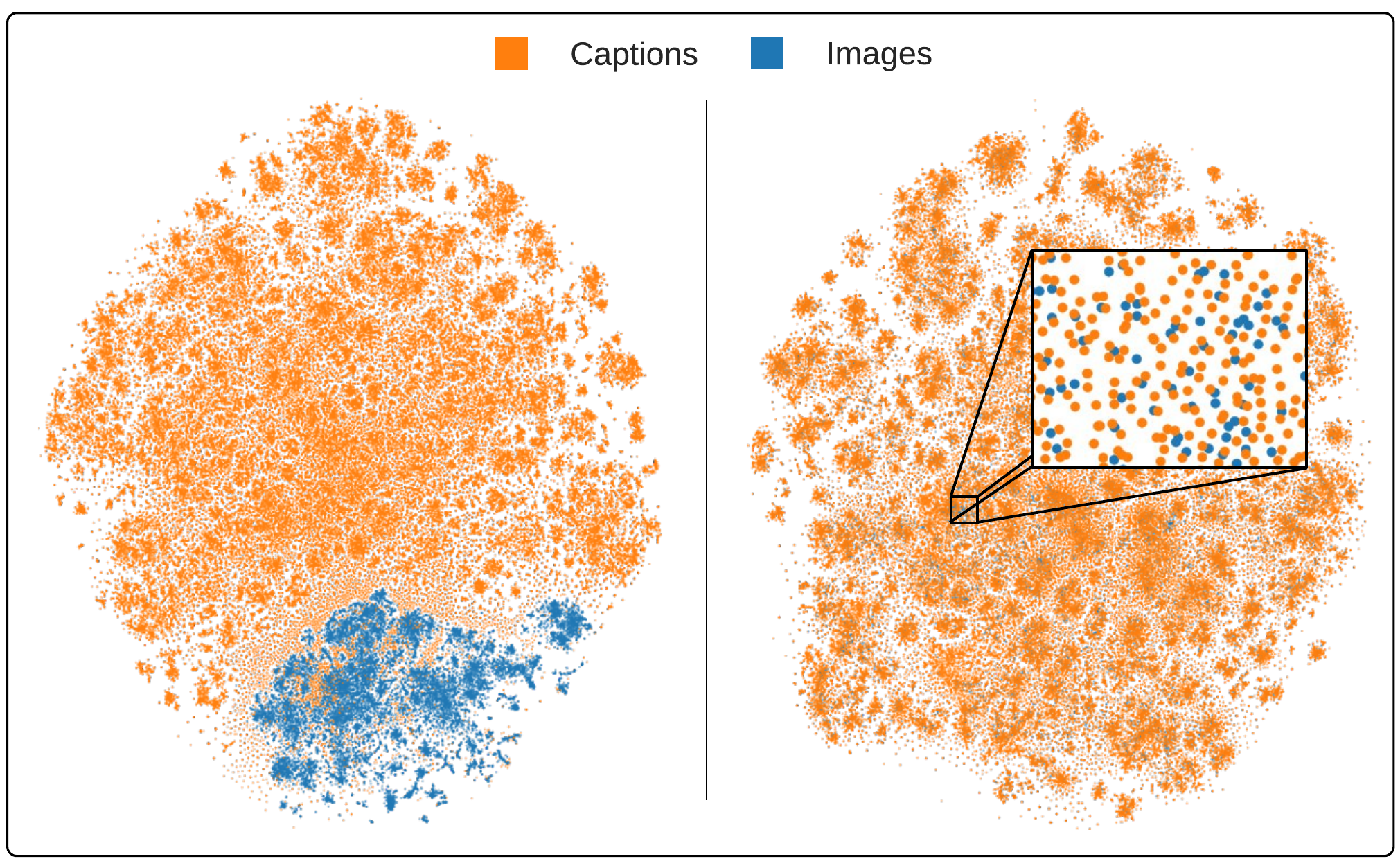}
    \end{center}
    \caption{T-SNE visualization of CLIP-embeddings before (left) and after (right) linear re-alignment on the Flickr30k dataset.}
    \label{fig:modality_gap}    
\end{wrapfigure}

\paragraph{Evaluation Metrics}
We report metrics commonly used for image captioning, such as CIDEr-D \citep{vedantam_cider_2015} and SPICE \citep{anderson_spice_2016}.\footnote{CIDEr-D and SPICE metrics are computed using the code from \url{https://github.com/tylin/coco-caption}.}
Further, we report CLIP-based metrics, CLIP-S and RefCLIP-S \citep{hessel_clipscore_2021}, CLIP+DN and CLIP+DN-Ref \citep{zhou_distribution_2023}, as well as our proposed metrics aCLIP-S and RefaCLIP-S.
We include error bars in the form of the standard error for all methods we trained ourselves to enable a thorough scientific comparison.
We do not report error bars for CIDEr-D and SPICE on VizWiz since the evaluation server does not provide them.
We highlight the best performing methods in boldface throughout the paper and consider two methods to be on-par when their standard errors overlap (68.2\% confidence intervals).

\paragraph{Results}
\cref{tab:caption_all} shows our results for MS-COCO and Flickr30k.
\OurMethod performs on-par or better than all competitors on our proposed metrics aCLIP-S and RefaCLIP-S on both datasets.
On Flickr30k, \OurMethod attains performance on-par with SmallCap in terms of CIDEr-D and SPICE even though \OurMethod trains about 1000 times faster with less trainable parameters (see \cref{tab:efficiency}).
While \OurMethod attains slightly lower scores on CIDEr-D and SPICE for MS-COCO, it performs on-par or better on CLIP-based metrics.
On VizWiz, \OurMethod outperforms competitors on most metrics.
Finally, on MSRVTT, \OurMethod significantly outperforms both ClipCap and SmallCap across all metrics.

We visualize the joint embedding space of the RN50$\times$64 CLIP encoder without applying our linear alignment for the Flickr30k training set (29K images and 145K captions) via t-SNE \citep{vandermaaten_visualizing_2008} in \cref{fig:modality_gap}, left.
We find that images and captions are not well aligned int the joint embedding space.
However, after applying our linear mapping the two modalities align very well, as shown in \cref{fig:modality_gap}, right.

\begin{wraptable}{r}{.5\textwidth}
\caption{Number of trainable parameters, training time, and inference time of \OurMethod compared to existing lightweight image captioning methods. Inference time is measured in seconds on a subset of 1000 images from the MS-COCO test set on an A100 GPU.}
\label{tab:efficiency}
\begin{center}
\begin{small}
\begin{sc}
\resizebox{.5\columnwidth}{!}{
    \begin{tabular}{l | c c}
    \toprule
    Method & $|\theta|$ & Training \\
    \midrule
    ClipCap & 42.8M & 6h (GTX1080) \\ %
    Prefix-Diffusion & 38.25M & n/a \\ %
    I-Tuning & 14M & n/a \\
    $\text{SmallCap}_{\text{d=4,Base}}$ & 1.8M & 8h(A100)  \\ %
    ReCap (ours) & \textbf{1.0M} & \textbf{20.3s $\pm$ 1.91 (CPU)} \\ %
    \bottomrule
\end{tabular}
}
\end{sc}
\end{small}
\end{center}
\vskip -0.1in
\end{wraptable}

\paragraph{Cross-domain Transfer}
Next, we investigate the cross-domain transfer of \OurMethod from MS-COCO to all other domains. 
We show results for three settings, where we use the same mapping trained on MS-COCO, but evaluate with different datastores, (i) the target datastore, (ii) the source datastore, and (iii) source and target datastores combined.
Here source always refers to MS-COCO data and target refers to one of Flickr30k, VizWiz, or MSRVTT.
For this line of experiments we only compare to SmallCap since it is the only existing lightweight captioning method that uses retrieval augmentation, and thus, accesses a datastore.
\cref{tab:mscoco_flickr_transfer} shows CIDEr-D and RefaCLIP-S scores if applicable, otherwise aCLIP-S scores, on all domains.
\OurMethod consistently outperforms SmallCap on aCLIP-S and RefaCLIP-S.
Further, \OurMethod consistently outperforms SmallCap when only retrieving from the target datastore, demonstrating improved transfer capabilities.
Combining data from both domains usually leads to a performance drop, which indicates that captions from the source domain interfere with the target domain.
Both methods are increasingly affected by the domain shift when using the datastore from the source domain.
However, \OurMethod still outperforms SmallCap on most metrics.
These results demonstrate improved transfer capabilities of \OurMethod by representing images in the form of text only.

\begin{table}
\caption{Transfer experiments for $\text{SmallCap}_{\text{d=4,Base}}$ and \OurMethod trained on MS-COCO and evaluated on the Flickr30k, VizWiz, and MSRVTT test sets. The datastore either contains data from the target domain, the source domain, or both of them combined. We report round mean and standard error of CIDEr-D and aCLIP-S/RefaCLIP-S for ReCap.}
\label{tab:mscoco_flickr_transfer}
\vskip 0.15in
\begin{center}
\begin{small}
\begin{sc}
\resizebox{\columnwidth}{!}{
    \begin{tabular}{l c c c c c c}
    \toprule
    Method & \multicolumn{2}{c}{Flickr30k} & \multicolumn{2}{c}{VizWiz} & \multicolumn{2}{c}{MSRVTT}  \\
    & CIDEr-D & RefaCLIP-S & CIDEr-D & aCLIP-S & CIDEr-D & RefaCLIP-S \\
    \midrule
    \multicolumn{7}{c}{Target Datastore}\\
    \midrule
    SmallCap & 59.3 $\pm$ 1.9 & 53.0 $\pm$ 0.2 & 51.0 $\pm$ 0.0 & 15.8 $\pm$ 0.1 & 19.5 $\pm$ 0.1 & 31.0 $\pm$ 0.0 \\
    ReCap (ours)  & \textbf{63.9 $\pm$ 1.9} & \textbf{54.6 $\pm$ 0.2} & \textbf{53.1 $\pm$ 0.0} & \textbf{24.4 $\pm$ 0.1} & \textbf{29.4 $\pm$ 0.1} & \textbf{34.3 $\pm$ 0.0} \\
    \midrule
    \multicolumn{7}{c}{Source + Target Datastore}\\
    \midrule
    SmallCap  & 50.4 $\pm$ 1.7 & 52.3 $\pm$ 0.2 & \textbf{51.0 $\pm$ 0.0} & 15.8 $\pm$ 0.1 & 19.5 $\pm$ 0.1 & 31.0 $\pm$ 0.0  \\ 
    ReCap (ours) & \textbf{58.9 $\pm$ 1.8} & \textbf{54.1 $\pm$ 0.2} & 49.8 $\pm$ 0.0 & \textbf{23.9 $\pm$ 0.1} & \textbf{25.5 $\pm$ 0.1} & \textbf{33.5 $\pm$ 0.0} \\
    \midrule
    \multicolumn{7}{c}{Source Datastore}\\
    \midrule
    SmallCap  & \textbf{48.9 $\pm$ 1.6} & 52.1 $\pm$ 0.2 & \textbf{36.1 $\pm$ 0.0} & 12.5 $\pm$ 0.1 & 16.5 $\pm$ 0.1 & 30.2 $\pm$ 0.0 \\
    ReCap (ours) & \textbf{48.5 $\pm$ 1.6} & \textbf{53.0 $\pm$ 0.2} & 28.6 $\pm$ 0.0 & \textbf{17.9 $\pm$ 0.1} & \textbf{17.5 $\pm$ 0.1} & \textbf{32.0 $\pm$ 0.0} \\
    \bottomrule
    \end{tabular}
}
\end{sc}
\end{small}
\end{center}
\vskip -0.1in
\end{table}

\subsection{Metrics for Image Captioning}
\label{sec:experiments_metric}

Following standard practice \citep{hessel_clipscore_2021,zhou_distribution_2023}, we evaluate our proposed metrics for image captioning by measuring their correlation with human rankings of candidate captions.

\paragraph{Datasets}
We use the Flickr8k-Expert (Flickr8k-E), Flickr8k-Crowdflower \citep[Flickr8k-CF]{hodosh_framing_2013}, and THumB datasets \citep{kasai_transparent_2022} . 
These datasets provide candidate captions along with human rankings for images of the test set of Flickr8k and MS-COCO.

\paragraph{Baselines}
We compare our metrics to the current state-of-the-art reference-based and reference-free metrics.
In the case of reference-free metrics, we compare to CLIP-score \citep{hessel_clipscore_2021}, and CLIP+DN \citep{zhou_distribution_2023}. 
We compare our reference-based metric to RefCLIPScore \citep{hessel_clipscore_2021}, CLIP+DN-Ref \citep{zhou_distribution_2023}, MID \citep{kim_mutual_2022}, and SoftSPICE \citep{li_factual_2023}, as well as rule-based metrics such as BLEU and CIDEr-D.
For all CLIP+DN variants (reference-based and reference-free) we estimate the mean of both modalities on the respective training dataset.
Further, we include a different vision encoder, namely SigLIP \citep{zhai_sigmoid_2023}, which has demonstrated improvements on cross-modal retrieval over CLIP variants.

\paragraph{Evaluation Metrics}
To quantify correlation with human judgement, we report Kendall's $\tau_c$ for Flickr8k-E and THumB, and Kendall's $\tau_b$ for Flickr8k-CF as done in prior work \citep{zhou_distribution_2023}.
The Kendall rank correlation coefficient measures the ordinal association between rankings by humans and the metric.

\paragraph{Results}
We report our results in \cref{tab:metric_eval}.
First, we note that aCLIP-S/RefaCLIP-S consistently outperform CLIP-S/RefCLIP-S from which they were derived. 
Remarkably, our linear alignment seems to be particularly effective for Flickr8K-E, while it sometimes even leads to a decreased score for Flickr8k-CF.
However, our linear alignment in combination with the SigLIP encoder reaches the highest score on average across all three datasets.
In the case of reference-based metrics, RefaSigLIP reaches the highest average correlation across all three datasets.
We show additional results for different CLIP vision encoders used for our metrics in \cref{sec:additional_results}.

\begin{table}
\caption{Correlation of different metrics with human judgement on the Flickr8k-E, Flickr8k-CF, and THumB datasets. We report Kendall's $\tau_c$ for every method. The standard error for $\tau$ depends only on the size of the test set and the number of captions per image and is equal for each method, i.e., 0.005 for Flickr-E, 0.003 for Flickr-CF, and 0.006 for THumB. Boldface indicates highest scores.}
\label{tab:metric_eval}
\vskip 0.15in
\begin{center}
\begin{small}
\begin{sc}
    \begin{tabular}{l| c c c |c}
    \toprule
    Method & Flickr8k-E & Flickr8k-CF & THumB & Avg \\
    \midrule
    \multicolumn{5}{c}{Reference-free} \\
    \midrule
    CLIP-S & 51.4 & 34.3 & 19.9 & 35.2 \\
    CLIP+DN & 54.0 & 35.2 & 23.3 & 37.5 \\
    SigLIP-B/16 & 47.0 & 42.3 & 23.0 & 37.4 \\
    SigLIP-L/16 & 43.9 & \textbf{45.6} & 25.4 & 38.3 \\
    aCLIP-S (ours) & 55.1 & 36.2 & 22.5 & 37.9 \\
    $\text{aSigLIP-B/16}$ (ours) & \textbf{55.5} & 36.7 & 24.3 & 38.8 \\
    $\text{aSigLIP-L/16}$ (ours) & 55.4 & 37.4 & \textbf{27.6} & \textbf{40.1} \\
    \midrule
    \multicolumn{5}{c}{Reference-based} \\
    \midrule
    BLEU@1 & 32.3 & 17.9 & 11.1 & 20.4 \\
    BLEU@4 & 30.8 & 16.9 & 6.9 & 18.2 \\
    CIDEr & 43.9 & 24.6 & 13.8 & 27.4 \\
    RefCLIP-S & 53.0 & 36.4 & 24.7 & 38.0 \\
    SoftSPICE & 54.2 & n/a & n/a & n/a\\
    MID & 54.9 & 37.3 & n/a & n/a \\
    CLIP+DN-Ref  & 55.0 & 37.0 & 27.1 & 39.7 \\
    RefSigLIP-B/16 & 47.2 & 42.5 & 24.7 & 38.1 \\
    RefSigLIP-L/16 & 43.9 & \textbf{45.8} & 27.4 & 39.0 \\
    RefaCLIP-S (ours) & 55.5 & 36.7 & 24.3 & 38.8 \\
    $\text{RefaSigLIP-B/16}$ (ours) & \textbf{55.8} & 37.2 & 26.0 & 39.7 \\
    $\text{RefaSigLIP-L/16}$ (ours) & \textbf{55.8} & 37.8 & \textbf{29.8} & \textbf{41.1} \\
    \bottomrule
\end{tabular}
\end{sc}
\end{small}
\end{center}
\end{table}

\subsection{Cross-modal Retrieval}
\label{sec:cross_modal_retrieval}
Since \OurMethod is based on retrieval augmentation, we conduct additional experiments to evaluate how captioning performance correlates with cross-modal retrieval performance.

\paragraph{Datasets}
We use the popular MS-COCO and Flickr30k cross-modal retrieval benchmarks, where the task is to retrieve a caption that belongs to an image (image$\rightarrow$text) and vice versa (text$\rightarrow$image).
In our setting we are particularly interested in the former, since image-to-text retrieval is an essential component of \OurMethod, however we report both to obtain a better understanding of the effect of the linear alignment.

\paragraph{Baselines}
We compare the most widely used CLIP model for retrieval (ViT-B/32) to a resnet-based variant (RN50$\times$64) and to their aligned versions via constrained ($\text{aCLIP}_\text{PR}$) and unconstrained ($\text{aCLIP}_\text{OLS}$) least squares mappings.
Further, we add a baseline that uses beta-procrustes which interpolates between the procrustes and an identity mapping. 
We also add two baselines that optimize the linear alignment iteratively ($\text{aCLIP}_\text{IT}$ and $\text{aCLIP}_\text{LFA}$), where $\text{aCLIP}_\text{IT}$ maximizes cosine similarity between image-caption pairs, and $\text{aCLIP}_\text{LFA}$ uses an adaptive re-ranking loss which has proven to be particularly effective in the few-shot classification setting \citep{ouali_black_2023}.

\paragraph{Results}
We evaluate all methods by measuring average recalls and cosine similarities and report our results in \cref{tab:retrieval_coco_flickr}.
Surprisingly, the best performing method in terms of image-to-text retrieval is the unaligned RN50$\times$64 CLIP encoder and also performs best across all publicly available CLIP encoders (see \cref{tab:retrieval_clip_ablation} in \cref{sec:additional_results}).
Aligned versions of CLIP do not improve image-to-text retrieval, but rather text-to-image retrieval.
While the performance on image-to-text retrieval decreases, we observe improved performance on image captioning (see \cref{tab:main_ablations} in \cref{sec:additional_results}).
An intuitive explanation for this is that in the image captioning setting there are not always clear boundaries between captions, i.e. classes. 
For example, an object appearing in one image might also appear in a different image.
Therefore the alignment process automatically increases the cosine similarity to all captions that semantically fit an image, leading to misclassifications that are heavily punished by the recall metric.
When considering cosine similarity between image and text embeddings though, we find that higher cosine similarity for the image-to-text direction also results in better captioning performance, as the best setting of \OurMethod is based on $\text{aCLIP}_\text{OLS}$. 
Further, we surmise that the discrepancy between recall and cosine similarity might be rooted in their continuity, i.e., that the recall metric is unable to capture moderate improvements due to its discontinuity \citep{schaeffer_emergent_2023}. 

\begin{table}[t]
\caption{Comparison of different CLIP vision encoders on cross-modal retrieval on MS-COCO and Flickr30k. We report average recalls and standard error for all methods, as well as average cosine similarity. All aCLIP variants use the RN50$\times$64 encoder. Boldface indicates highest average scores.}
\label{tab:retrieval_coco_flickr}
\begin{center}
\begin{small}
\begin{sc}
\begin{tabular}{l | c c c | c c c }
    \toprule
    & \multicolumn{6}{c}{MS-COCO} \\
    & \multicolumn{3}{c|}{Image $\rightarrow$ Text} & \multicolumn{3}{c}{Text $\rightarrow$ Image} \\
    \midrule
    Method & R@1 & R@5 & cos($\theta$) & R@1 &  R@5 & cos($\theta$) \\
    \midrule
    $\text{CLIP}_{\text{RN50x64}}$ & \textbf{60.7 $\pm$ 0.7} & \textbf{82.2 $\pm$ 0.5} & 0.297 & 34.3 $\pm$ 0.5 & 59.5 $\pm$ 0.5 & 0.288 \\
    $\text{CLIP}_{\text{ViT-B/32}}$ & 52.3 $\pm$ 0.7 & 76.0 $\pm$ 0.6 & 0.343 & 30.2 $\pm$ 0.5 & 55.1 $\pm$ 0.5 & 0.335 \\
    $\text{aCLIP}_\text{LFA}$ & 57.1 $\pm$ 0.7 & 80.0 $\pm$ 0.6 & 0.318 & 40.1 $\pm$ 0.5 & 65.0 $\pm$ 0.5 & 0.301 \\
    $\text{aCLIP}_{\text{PR}}$ & 45.3 $\pm$ 0.7 & 69.7 $\pm$ 0.7 & 0.512 & 35.4 $\pm$ 0.5 & 59.4 $\pm$ 0.5 & 0.477 \\
    $\text{aCLIP}_{\beta\text{-PR}}$ & 55.8 $\pm$ 0.7 & 79.8 $\pm$ 0.6 & 0.558 & 37.5 $\pm$ 0.5 & 62.2 $\pm$ 0.5 & 0.292 \\
    $\text{aCLIP}_{\text{OLS}}$ & 33.3 $\pm$ 0.7 & 59.2 $\pm$ 0.7 & \textbf{0.699} &  \textbf{41.5 $\pm$ 0.5} & \textbf{66.9 $\pm$ 0.5} & \textbf{0.619} \\
    $\text{aCLIP}_{\text{IT}}$ & 33.1 $\pm$ 0.7 & 60.3 $\pm$ 0.7 & 0.320 &  31.6 $\pm$ 0.5 & 57.1 $\pm$ 0.5 & 0.288 \\
    \midrule
    & \multicolumn{6}{c}{Flickr30k} \\\midrule
    $\text{CLIP}_{\text{RN50x64}}$ & \textbf{88.5 $\pm$ 1.0} & \textbf{98.3 $\pm$ 0.4} & 0.303 &  69.1 $\pm$ 1.0 & 90.7 $\pm$ 0.6 & 0.282 \\
    $\text{CLIP}_{\text{ViT-B/32}}$ & 79.8 $\pm$ 1.2 & 96.3 $\pm$ 0.6 & 0.347 & 59.3 $\pm$ 1.1 & 83.7 $\pm$ 0.8 & 0.330 \\
    $\text{aCLIP}_\text{LFA}$ & 79.2 $\pm$ 1.3 & 95.5 $\pm$ 0.7 &  0.457 & 67.5 $\pm$ 1.0 & 89.6 $\pm$ 0.6 & \textbf{0.675} \\
    $\text{aCLIP}_{\text{PR}}$ & 78.5 $\pm$ 1.3 & 95.1 $\pm$ 0.7 & 0.460 & 67.0 $\pm$ 1.0 & 89.2 $\pm$ 0.6 & 0.403 \\
    $\text{aCLIP}_{\beta\text{-PR}}$ & 85.7 $\pm$ 1.1 & 97.5 $\pm$ 0.5 & 0.403 & \textbf{72.6 $\pm$ 1.0} & \textbf{92.5 $\pm$ 0.5} & 0.356 \\
    $\text{aCLIP}_{\text{OLS}}$ & 73.6 $\pm$ 1.4 & 95.0 $\pm$ 0.7 & \textbf{0.624} & 70.6 $\pm$ 1.0 & 90.6 $\pm$ 0.6 & 0.547 \\
    $\text{aCLIP}_{\text{IT}}$ & 67.3 $\pm$ 1.5 & 90.5 $\pm$ 0.9 & 0.308 &  62.8 $\pm$ 1.0 & 86.1 $\pm$ 0.7 & 0.268 \\
    \bottomrule
\end{tabular}
\end{sc}
\end{small}
\end{center}
\vskip -0.1in
\end{table}

\section{Related Work}
\label{sec:related_work}

\paragraph{Linear Alignment}

The idea of linearly aligning embedding spaces is a well studied problem in the field of bilinguality \citep{minixhofer_wechsel_2022,artetxe_learning_2016}, geometrical alignment \citep{leordeanu_spectra_2005,fischler_random_1981,liu_sift_2008}, and vision for zero-shot learning \citep{akata_label_2013,akata_evaluation_2015,frome_devise_2013,bernadino_embarrasingly_2015}.
Similar to our approach, \citet{ouali_black_2023} use the procrustes method to align features of CLIP with embedded class labels for few-shot classification.
Other works sidestep the prevalent mis-aligned embedding space by training a decoder solely in the text space of CLIP \citep{li_decap_2023,nukrai_textonly_2022,yu_multimodal_2022,wang_from_2023,gu_i_2022}.
At test time, however, these approaches receive images as input and, thus, still suffer from the prevalent mis-alignment.
Other approaches adapt the pretraining objective in order to achieve a better alignment in the joint embedding space \citep{furst_cloob_2022,goel_cyclip_2022,humer_understanding_2023}.
However, none of these models are available at the same scale as CLIP.

\paragraph{Retrieval Augmentation}

The idea of retrieval augmentation has been explored in the realm of language modeling \citep{khandelwal_generalization_2020,guu_retrieval_2020,borgeaud_improving_2022}, language generation conditioned on images \citep{hu_reveal_2023,yang_revilm_2023,yasunaga_retrieval_2023}, and reinforcement learning \citep{humphreys_large_2022,goyal_retrieval_2022}.
In the realm of image captioning, \citet{ramos_smallcap_2023} leverages retrieval augmentation to reduce the required number of trainable parameters.
\citet{ramos_lmcap_2023} extends this idea to multilingual datastores, which enables generation in a certain target language.
\OurMethod also relies on retrieval augmentation, but is much more efficient in terms of training while yielding competitive or even better results.

\paragraph{Lightweight Image Captioning}

Lightweight captioning aims at reducing the training footpring for image captioning models.
One line of work is based on knowledge distillation \citep{hinton_distilling_2015} and assumes access to teacher captioning models that are distilled into much smaller scale models \citep{wang_efficient_2023,fang_compressing_2021,wang_minivlm_2020}.
Another line of works leverage parameter-efficient fine-tuning methods to merge visual knowledge into generative LMs via adapter layers \citep{eichenberg_magma_2022,zhang_llamaadapter_2023,gao_llamaadapter_2023}, cross-attention modules \citep{lup_ituning_2023,ramos_smallcap_2023}, or a mapping network between embedding spaces \citep{mokady_clipcap_2021,merullo_linearly_2022}.
Finally, while being lightweight, \citet{kuo_haav_2023} relies on a two-stage training procedure that includes fine-tuning via reinforcement learning \citep{li_oscar_2020,vinyals_show_2015,cornia_meshed_2020}. 
In contrast to \OurMethod, these methods require end-to-end training.

\section{Conclusion} 
In this work, we propose to leverage linear alignment techniques that can be computed in closed form for two use cases, image captioning and caption evaluation. 
We introduce \OurMethod, an efficient retrieval-augmented image-captioning method, which is based on linear alignment and requires substantially less training time than other lightweight image-captioning methods. 
We also introduce aCLIP-S and RefaCLIP-S, two new caption evaluation metrics that use linear alignment to adapt CLIP-S and RefCLIP-S, respectively, to a downstream dataset.
Since the evolution of the field is guided by the metrics that it uses, we envision that, by introducing metrics that correlate stronger with human perception than their predecessors, this work facilitates image-captioning research. 
We evaluate \OurMethod using rule-based metrics and find its performance to be similar to prior lightweight methods at substantially lower training costs. 
In terms of CLIP-based metrics, though, we find that \OurMethod outperforms competitors on all tasks thus improving the efficiency of lightweight image captioning systems on both ends. 
Finally, we demonstrate that \OurMethod improves transfer to different domains compared to existing lightweight retrieval-augmented methods demonstrating that \OurMethod generalizes well beyond the downstream task distribution.

\section*{Acknowledgements}
We are grateful to Wei Lin for his support, fruitful discussions and corrections.

The ELLIS Unit Linz, the LIT AI Lab, the Institute for Machine Learning, are supported by the Federal State Upper Austria. We thank the projects AI-MOTION (LIT-2018-6-YOU-212), DeepFlood (LIT-2019-8-YOU-213), Medical Cognitive Computing Center (MC3), INCONTROL-RL (FFG-881064), PRIMAL (FFG-873979), S3AI (FFG-872172), DL for GranularFlow (FFG-871302), EPILEPSIA (FFG-892171), AIRI FG 9-N (FWF-36284, FWF-36235), AI4GreenHeatingGrids(FFG- 899943), INTEGRATE (FFG-892418), ELISE (H2020-ICT-2019-3 ID: 951847), Stars4Waters (HORIZON-CL6-2021-CLIMATE-01-01). We thank Audi.JKU Deep Learning Center, TGW LOGISTICS GROUP GMBH, Silicon Austria Labs (SAL), University SAL Labs initiative, FILL Gesellschaft mbH, Anyline GmbH, Google, ZF Friedrichshafen AG, Robert Bosch GmbH, UCB Biopharma SRL, Merck Healthcare KGaA, Verbund AG, GLS (Univ. Waterloo) Software Competence Center Hagenberg GmbH, T\"{U}V Austria, Frauscher Sensonic, Borealis AG, TRUMPF and the NVIDIA Corporation.

\bibliographystyle{neurips2024}
\bibliography{neurips2024}

\appendix

\section*{Supplementary Material}

First, we elaborate on the potential societal impact of our work. 
Further, we provide the source code to reproduce all our experiments in \cref{app:source_code}.
To provide further insights into our method \OurMethod, we provide additional results on cross-modal retrieval, ablation studies, effect of different data sources, our DAL, and our evaluation as image captioning metric in \cref{sec:additional_results}.
Further, we provide more qualitative analysis on retrieved captions after the linear alignment and the effect of synthetic captions in \cref{app:qualitative_analysis}.
\cref{app:motivation} gives a rigorous theoretical intuition on the motivation of our linear alignment.
Finally, \cref{app:hyperparams} elaborates on the different hyperparameters we searched, including the retrieval parameter $k$, the decoding strategy, different vision encoders, generative language models, etc.

\section{Impact Statement}
This paper presents work whose goal is to advance the field of Machine Learning. 
There are many potential societal consequences of our work, on the forefront of such is the potential generation of misinformation or harmful content.
The method proposed in this work is based on CLIP and FLAN-T5. 
Any biases inherent to these models might also affect our method and, therefore, be present in captions produced by our method. 
A potential scientific conflict might arise from the usage of CLIP for both caption generation and evaluation as this potentially promotes CLIP's weaknesses and biases over iterations of model design and evaluation. 
In our experiments we found that different CLIP encoders lead to different performance on caption retrieval versus caption evaluation and also use different encoders for these two roles.
The method proposed in this paper aims at reducing the computational cost of model training for image captioning compared to conventional methods, thus reducing the energy footprint of image captioning systems while still providing state-of-the-art performance.

\section{Source Code}
\label{app:source_code}

To facilitate reproducibility of our findings, we will make the code publicly available upon acceptance.

\section{Baseline Implementation Details}
\label{sec:app_impl_details}

\paragraph{ClipCap}
We use the implementation of ClipCap available at \url{https://github.com/rmokady/CLIP_prefix_caption}.
We use the default parameters as specified in the codebase.
As a vision encoder we select the CLIP RN50x4 model since it is the only one that matches the reported parameter count in \citet{mokady_clipcap_2021}.
We use a batch size fof 40 and dump a checkpoint every 10,000 update steps.
After training we evaluate all checkpoints on the evaluation set and choose the one that reaches the highest score in terms of CIDEr-D score for evaluation on the test set.

\paragraph{SmallCap}
We train our SmallCap models with the code at \url{https://github.com/RitaRamo/smallcap}.
We use the same hyperparameters as \citet{ramos_smallcap_2023} and save checkpoints after every epoch.
As for ClipCap, we evaluate all checkpoints on the validation sets and select the one that reaches the highest CIDEr-D score.
Finally, we evaluate the selected checkpoint on the test set.

\section{Additional Results}
\label{sec:additional_results}

\paragraph{Cross-modal retrieval}

We evaluate all publicly available CLIP vision encoders on cross-modal retrieval on the MS-COCO and Flickr30k datasets.
We report average recalls and standard error in \cref{tab:retrieval_coco_flickr}.
We find that larger models improve retrieval performance and, perhaps surprisingly, the RN50$\times$64 encoder outperforms the largest ViT variant in four out of 6 categories when considering image to text retrieval on MS-COCO and Flickr30k.
Since \OurMethod is based on image to text retrieval we select RN50$\times$64 as our retrieval model.

\begin{table}[p]
\caption{Comparison of different CLIP vision encoders on the cross-modal retrieval task on MS-COCO and Flickr30k. We report average recalls and standard error for all publicly available CLIP vision encoders. Boldface indicates highest average scores.}
\label{tab:retrieval_clip_ablation}
\vskip 0.15in
\begin{center}
\begin{small}
\begin{sc}
\begin{tabular}{l | c c c | c c c }
    \toprule
    & \multicolumn{6}{c}{MS-COCO} \\
    & \multicolumn{3}{c|}{Image $\rightarrow$ Text} & \multicolumn{3}{c}{Text $\rightarrow$ Image} \\
    \midrule
    Method & R@1 & R@5 & R@10 & R@1 &  R@5 & R@10 \\
    \midrule
    $\text{CLIP}_{\text{RN50}}$ & 50.2 $\pm$ 0.7 & 74.9 $\pm$ 0.6 & 83.3 $\pm$ 0.5 & 28.4 $\pm$ 0.5 & 52.6 $\pm$ 0.5 & 64.2 $\pm$ 0.5 \\
    $\text{CLIP}_{\text{RN50x4}}$ & 52.2 $\pm$ 0.7 & 75.9 $\pm$ 0.6 & 67.5 $\pm$ 0.5 & 31.3 $\pm$ 0.5 & 55.7 $\pm$ 0.5 & 66.5 $\pm$ 0.5 \\
    $\text{CLIP}_{\text{RN50x16}}$ & 53.6 $\pm$ 0.7 & 77.9 $\pm$ 0.6 & 85.8 $\pm$ 0.5 & 33.2  $\pm$ 0.5 & 57.0 $\pm$ 0.5 & 67.5 $\pm$ 0.5 \\
    $\text{CLIP}_{\text{RN50x64}}$ & \textbf{60.7 $\pm$ 0.7} & \textbf{82.2 $\pm$ 0.5} & \textbf{88.5 $\pm$ 0.5} & 34.3 $\pm$ 0.5 & 59.5 $\pm$ 0.5 & 69.9 $\pm$ 0.5 \\
    $\text{CLIP}_{\text{ViT-B/32}}$ & 52.3 $\pm$ 0.7 & 76.0 $\pm$ 0.6 & 84.4 $\pm$ 0.5 & 30.2 $\pm$ 0.5 & 55.1 $\pm$ 0.5 & 66.4 $\pm$ 0.5\\
    $\text{CLIP}_{\text{ViT-B/16}}$ & 52.6 $\pm$ 0.7 & 76.9 $\pm$ 0.6 & 85.0 $\pm$ 0.5 & 32.9 $\pm$ 0.5 & 57.7 $\pm$ 0.5 & 68.1 $\pm$ 0.5 \\
    $\text{CLIP}_{\text{ViT-L/14}}$ & 57.0 $\pm$ 0.7 & 80.5 $\pm$ 0.6 & 86.9 $\pm$ 0.5 & 36.1 $\pm$ 0.5 & 60.3 $\pm$ 0.5 & 70.3 $\pm$ 0.5 \\
    $\text{CLIP}_{\text{ViT-L/14@336px}}$ & 58.5 $\pm$ 0.7 & 81.3 $\pm$ 0.6 & 88.1 $\pm$ 0.5 & 35.9 $\pm$ 0.5 & 60.4 $\pm$ 0.5 & 70.5 $\pm$ 0.5 \\
    \midrule
    & \multicolumn{6}{c}{Flickr30k} \\\midrule
    $\text{CLIP}_{\text{RN50}}$ & 80.8 $\pm$ 1.3 & 95.4 $\pm$ 0.7 & 97.8 $\pm$ 0.5 &  57.9 $\pm$ 1.1 & 83.1 $\pm$ 0.8 & 89.8 $\pm$ 0.6\\
    $\text{CLIP}_{\text{RN101}}$ & 79.2 $\pm$ 1.3 & 94.8 $\pm$ 0.7 & 97.8 $\pm$ 0.5 &  57.5 $\pm$ 1.1 & 81.9 $\pm$ 0.8 & 88.6 $\pm$ 0.7\\
    $\text{CLIP}_{\text{RN50x4}}$ & 83.0 $\pm$ 1.2 & 95.9 $\pm$ 0.6 & 98.2 $\pm$ 0.4 & 61.6 $\pm$ 1.1 & 84.7 $\pm$ 0.8 & 90.1 $\pm$ 0.6 \\
    $\text{CLIP}_{\text{RN50x16}}$ & 84.2 $\pm$ 1.2 & 97.0 $\pm$ 0.5 & 99.2 $\pm$ 0.3 & 64.5 $\pm$ 1.1 & 85.9 $\pm$ 0.7 & 91.5 $\pm$ 0.6 \\
    $\text{CLIP}_{\text{RN50x64}}$ & \textbf{88.5 $\pm$ 1.0} & 98.3 $\pm$ 0.4 & 99.4 $\pm$ 0.2 &  69.1 $\pm$ 1.0 & \textbf{90.7 $\pm$ 0.6} & \textbf{95.0 $\pm$ 0.4} \\
    $\text{CLIP}_{\text{ViT-B/32}}$ & 79.8 $\pm$ 1.2 & 96.3 $\pm$ 0.6 & 98.6 $\pm$ 0.4 & 59.3 $\pm$ 1.1 & 83.7 $\pm$ 0.8 & 90.3 $\pm$ 0.6 \\
    $\text{CLIP}_{\text{ViT-B/16}}$ & 83.0 $\pm$ 1.2 & 96.3 $\pm$ 0.6 & 99.3 $\pm$ 0.3 & 63.0 $\pm$ 1.1 & 85.9 $\pm$ 0.7 & 91.8 $\pm$ 0.6 \\
    $\text{CLIP}_{\text{ViT-L/14}}$ & 85.7 $\pm$ 1.1 & 98.3 $\pm$ 0.4 & 99.3 $\pm$ 0.3 & 64.8 $\pm$ 1.1 & 87.3 $\pm$ 0.7 & 92.4 $\pm$ 0.5\\
    $\text{CLIP}_{\text{ViT-L/14@336px}}$ & 88.5 $\pm$ 1.0 & \textbf{99.3 $\pm$ 0.3} & \textbf{99.6 $\pm$ 0.2} & 67.0 $\pm$ 1.0 & 88.7 $\pm$ 0.7 & 93.4 $\pm$ 0.5\\
    \bottomrule
\end{tabular}
\end{sc}
\end{small}
\end{center}
\vskip -0.1in
\end{table}

\paragraph{Impact of Linear Alignment}

We conduct an ablation study where we assess the effect of the linear alignment.
To this end, we evaluate a setting where we do not use our linear alignment, which we call $\text{\OurMethod}_{\text{ZS}}$, where ZS stands for zero-shot, since it does not require any training.
Further, we distinguish between two types of linear alignment, (i) constrained using orthogonal procrustes (PR), and (ii), unconstrained using ordinary least squares (OLS).
Results on the MS-COCO test set are shown in \cref{tab:main_ablations}.
We observe a substantial performance drop on all metrics for $\text{\OurMethod}_{\text{ZS}}$, showcasing the effectiveness of our linear alignment.
The best performing method in terms of CIDEr-D and SPICE is $\text{\OurMethod}_{\text{OLS}}$, since the unconstrained mapping leads to a stronger alignment with reference captions.
The best performance on our learning-based metrics is achieved by \OurMethod.
On one hand we observe the trend that on OLS alignment achieves a better trade-off between rule-based and our learning-based metrics.
The PR alignment on the other hand diverges more from reference captions and attains the best performance on our learning-based metrics.
Further, as we show in \cref{tab:metric_eval}, the PR alignment leads to higher correlation with human judgement.

Thus, we recommend the following criterion for when to deploy which optimization scheme:
\begin{itemize}
    \item For retrieval-augmented caption generation, use OLS
    \item For caption evaluation use PR 
\end{itemize}

\begin{table}[p]
\caption{Ablation study for different methods to compute our linear alignment on the MS-COCO test set. We compare unimodal retrieval (UM), the constrained mapping (PR), unconstrained mapping (OLS), and using no mapping at all (ZS). We report mean and standard error for all settings.}
\label{tab:main_ablations}
\vskip 0.15in
\begin{center}
\begin{small}
\begin{sc}
    \begin{tabular}{l c c c c}
    \toprule
    Method & CIDEr-D & SPICE & aCLIP & RefaCLIP-S\\
    \midrule
    $\text{ReCap}_{\text{UM}}$ & 81.9 $\pm$ 0.9 & 16.6 $\pm$ 0.1 & 46.1 $\pm$ 0.1 & 56.0 $\pm$ 0.1 \\
    $\text{ReCap}_{\text{ZS}}$ & 92.2 $\pm$ 0.9 & 19.3 $\pm$ 0.1 & 46.1 $\pm$ 0.1 &  57.2 $\pm$ 0.1 \\
    $\text{ReCap}_{\text{IT}}$ & 91.0 $\pm$ 0.9  & 18.7 $\pm$ 0.1 & 46.1 $\pm$ 0.1 &  57.2 $\pm$ 0.1 \\
    $\text{ReCap}_{\beta\text{-PR}}$ & 94.8 $\pm$ 1.0 & 19.4 $\pm$ 0.1 & 46.1 $\pm$ 0.1 &  57.6 $\pm$ 0.1 \\
    $\text{ReCap}_\text{LFA}$ & 107.5 $\pm$ 1.0 & 20.6 $\pm$ 0.1 & 46.1 $\pm$ 0.1 &  57.8 $\pm$ 0.1\\
    $\text{ReCap}_{\text{PR}}$ & 104.9 $\pm$ 1.0 & 20.4 $\pm$ 0.1 & 46.1 $\pm$ 0.1 & 57.9 $\pm$ 0.1 \\
    $\text{ReCap}_{\text{OLS}}$ & \textbf{108.3 $\pm$ 1.0} & \textbf{21.2 $\pm$ 0.1} &  46.1 $\pm$ 0.1 & \textbf{58.0 $\pm$ 0.1} \\
    \bottomrule
\end{tabular}
\end{sc}
\end{small}
\end{center}
\vskip -0.1in
\end{table}

\paragraph{Effect of different data sources}

We conduct another line of experiments where we investigate the effect of additional data sources in the datastore. 
To this end, we use \OurMethod aligned to MS-COCO data and add data from Flickr30k, VizWiz, MSRVTT, and synthetic captions from our DAL to the datastore.
In \cref{tab:coco_additional_data_sources} we report CIDEr-D, SPICE, aCLIP, and RefaCLIP for all settings.
Generally, we observe that our synthetic captions have the most impact on captioning performance on our aCLIP-S and RefaCLIP-S metrics.
For the remaining metrics we do not observe a significant difference independent of the added data source.
This means that even though the datastore grows, there is not much difference in the captions that are provided to the LM in the prompt, i.e. even though captions are added, they are never retrieved.
This is different for synthetic captions though, and thus, illustrates the potential utility of high quality synthetic captions.

\begin{table}
\caption{Training-free use of additional data sources on the MS-COCO (CO) test set for $\text{\OurMethod}_{\text{OLS}}$. Additional data sources include captions from Flickr30k (F30), VizWiz (VW), MSRVTT (MV), and synthetic captions (SC) from DAL.  We report mean and standard error, if it exceeds a threshold of 1e-4, for all metrics.}
\label{tab:coco_additional_data_sources}
\vskip 0.15in
\begin{center}
\begin{small}
\begin{sc}
    \begin{tabular}{l c c c c}
    \toprule
    Datastore & CIDEr-D & SPICE & aCLIP-S & RefaCLIP-S \\
    \midrule
    $\text{CO}$ & \textbf{108.3 $\pm$ 1.0} &  21.2 $\pm$ 0.1 & 46.1 $\pm$ 0.1 & 58.0 $\pm$ 0.1 \\
    $\text{CO}+\text{F30}$ & 107.9 $\pm$ 1.0 & 21.1 $\pm$ 0.1 & 46.1 $\pm$ 0.1 & 58.0 $\pm$ 0.1 \\
    $\text{CO}+\text{F30}+\text{VW}$ & 108.0 $\pm$ 1.0 & 21.2 $\pm$ 0.1 &  46.1 $\pm$ 0.1 & 58.0 $\pm$ 0.1\\
    $\text{CO}+\text{F30}+\text{MV}$ & 108.2 $\pm$ 1.0 & 21.2 $\pm$ 0.1 &  46.1 $\pm$ 0.1 & 58.0 $\pm$ 0.1 \\
    $\text{CO}+\text{F30}+\text{VW}+\text{MV}$ & 108.2 $\pm$ 1.0 & 21.2 $\pm$ 0.1 &  46.1 $\pm$ 0.1 & 58.0 $\pm$ 0.1 \\
    \bottomrule
    \end{tabular}
\end{sc}
\end{small}
\end{center}
\vskip -0.1in
\end{table}

\paragraph{Datastore-augmentation Loop}

In this section we elaborate on preliminary results on adding synthetic captions generated by the LM to the retrieval datastore.
We aim to add synthetic captions of high quality to the datastore, such that the over-all prediction quality of \OurMethod improves. 
To measure the quality of synthetic captions we assume access to a metric $\mu : \cT \times \cT \to \dR$.\footnote{We use notation for a reference-based metric. However, DAL works just as well with a reference-free metric.}
We start by evaluating \OurMethod on the validation set and compute the average metric $\bar \mu$, which provides us with an estimate of the quality of generated captions.
Next, we iterate over images from $\cD_{\text{Train}}$ and create synthetic captions via \OurMethod.
After caption generation we compute $\mu(\cdot,\cdot)$ for every synthetic caption candidate and add only those to the datastore for which the score exceeds $\bar \mu$.
Then we evaluate on $\cD_{\text{val}}$ again and update $\bar \mu$.
We repeat this process for a fixed number of iterations.
\cref{alg:self_improvement} shows the pseudocode for our proposed DAL.

\begin{algorithm*}[h]
   \caption{Datastore-augmentation Loop via Synthetic Captions}
   \label{alg:self_improvement}
\begin{algorithmic}
   \Require caption metric $\mu(\cdot, \cdot)$, CLIP vision encoder $\phi(\cdot)$, CLIP text encoder $\psi(\cdot)$, batched nucleus sampling from language model $\texttt{LM}(\cdot, \cdot)$, training set $\cD_{\text{Train}}$, validation set $\cD_{\text{Val}}$, prompt $\Bp$, hyperparameters $k, l, m \in \dN$
   \State
   \State $\BW \gets \texttt{fit\_linear}\{(\phi(\Bx), \psi(\Bc)) \mid (\Bx, \Bc) \in \cD_{\text{Train}}\}$ \Comment{Re-align CLIP for downstream data; cf.\ Eq.\ \eqref{eq:procrustes}}
   \State $\cC \gets \{ \Bc \mid (\Bx, \Bc) \in \cD_{\text{Train}} \}$ \Comment{Initialize datastore with training captions}
   \State
   \Function{ReCap}{$\Bx, \BW, \cC$} 
        \State $\cK \gets \operatorname*{arg\,max}_{\Bc \in \cC}^k \operatorname{cossim}(\psi(\Bc), \BW \phi(\Bx))$ \Comment{Select top-$k$ captions for $\Bx$; cf.\ Eq.\ \eqref{eq:topk}}
        \State $\Bq \gets \texttt{concat}(\{\Bp\} \cup \cK)$ \Comment{Combine top-$k$ captions into one prompt}
        \State $\cS \gets \texttt{LM}(\Bq, l)$ \Comment{Sample $l$ responses of LM via nucleus sampling}
        \State \Return $ \operatorname*{arg\,max}_{\Bs \in \cS}\operatorname{cossim}(\psi(\Bs), \BW \phi(\Bx))$ \Comment{Return the response that fits $\Bx$ best; cf.\ Eq.\ \eqref{eq:filtering}}
    \EndFunction
   \State
   \For{$i \in \{1, \dots, m\}$}
        \State $\bar \mu \gets \frac{1}{\mid \cD_{\text{Val}} \mid} \sum_{(\Bx, \Bc) \in \cD_{\text{Val}}} \mu($\Call{ReCap}{$\Bx, \BW, \cC$}$, \Bc)$ \Comment{Compute average validation score}
        \State $\cC \gets \cC \cup \{ \Bc' \mid \Bc' = $ \Call{ReCap}{$\Bx, \BW, \cC$} $ \land\,\mu(\Bc', \Bc) > \bar \mu \land (\Bx, \Bc) \in \cD_{\text{Train}}\}$ \Comment{Add synthetic captions}
    \EndFor
\end{algorithmic}
\end{algorithm*}

We run our DAL for $m=5$ iterations and instantiate $\mu(\cdot,\cdot)$ with CIDEr-D, SPICE, aCLIP-S, and RefaCLIP-S to filter the synthetic captions.
If more than one synthetic caption exceeds the threshold $\Bar{\mu}$, we only take the highest scoring one.
After each round of augmentation we search over the hyperparameter $k$ that yields the highest average score $\Bar{\mu}(\cdot,\cdot)$ on the validation set.
Finally, we evaluate the datastore with the found $k$ on the test set to measure final performance.

\begin{figure}
    \centering
    \includegraphics[width=.7\textwidth]{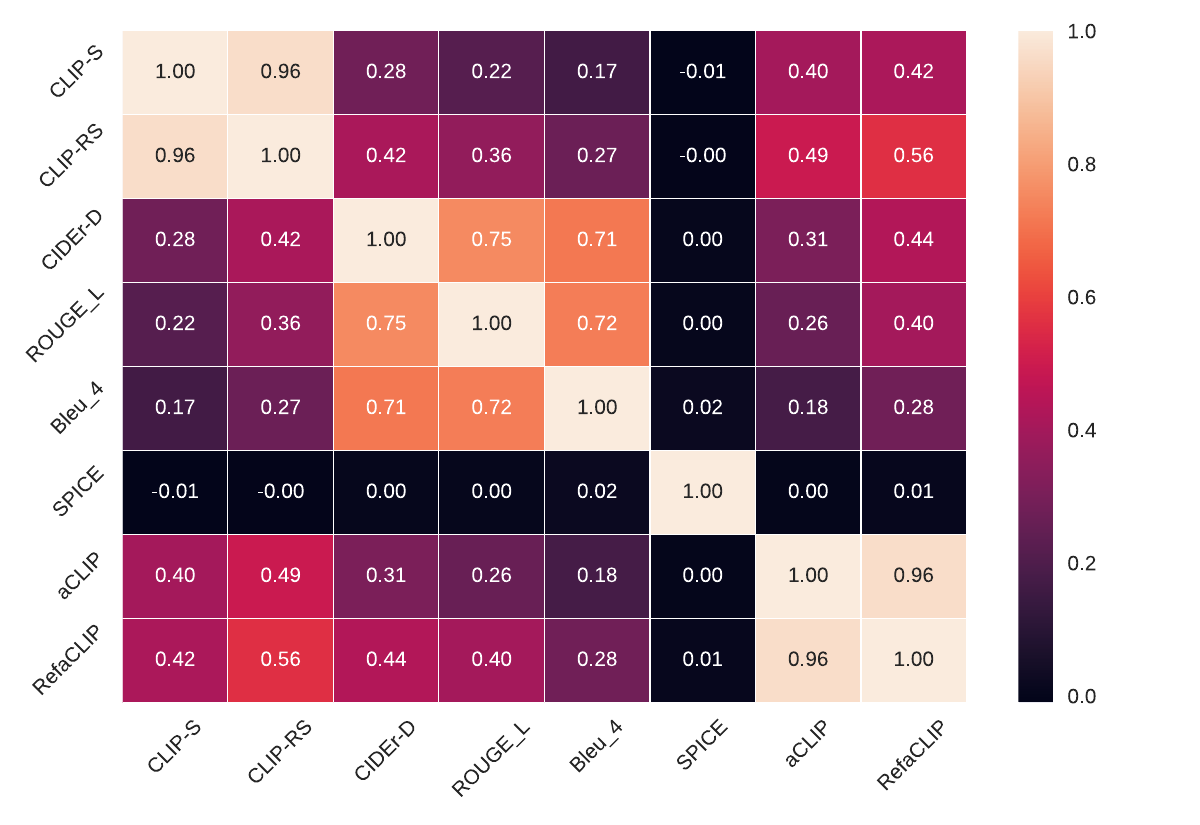}
    \vskip -1em
    \caption{Pearson correlation between commonly used image captioning metrics for captions generated via $\text{ReCap}$ on the MS-COCO test set.}
    \label{fig:metrics_correlation}   
\end{figure}

We apply DAL to \OurMethod for both MS-COCO and Flickr30k datasets.
Per iteration, DAL adds on average $42320$ and $35288$ synthetic captions to the datastore for MS-COCO and Flickr30k, respectively. 
This corresponds to 7\% and 24\% of the original datastore sizes, respectively.
We find that the selection of the metric for filtering synthetic captions in DAL is non-trivial.
Filtering with respect to one metric usually leads to performance improvements on this very metric.
This is due to a rather low correlation between metrics as we show in \cref{fig:metrics_correlation}.
Metrics, such as BLEU, ROUGE-L and CIDEr-D correlate strongly with each other.
This is due to the fact, that they all rely on n-gram based matching to reference captions.
Further, CLIP-S and CLIP-RS correlate strongly with each other, since they are both based on cosine similarity by CLIP.
The same is true for aCLIP-S, and RefaCLIP-S, which are both based on cosine similarity of our aligned CLIP.
However, aCLIP-S and RefaCLIP-S both correlate stronger with n-gram based metrics than CLIP-S and RefCLIP-S due to the alignment to reference captions.
Interestingly, SPICE is entirely decorrelated to all other metrics, since it is based on semantic scene graphs.
This indicates that some of these metrics evaluate different aspects of human judgement, thus, optimizing for one metric does not necessarily lead to improvement in any other metric.
Interestingly, the correlation between our aCLIP-S metrics and CLIP-S metrics is, perhaps, lower than one might expect.
This indicates that our proposed metrics behave differently to CLIP-S and are more geared toward the human annotated references.

\begin{figure}
    \centering
    \includegraphics[width=\textwidth]{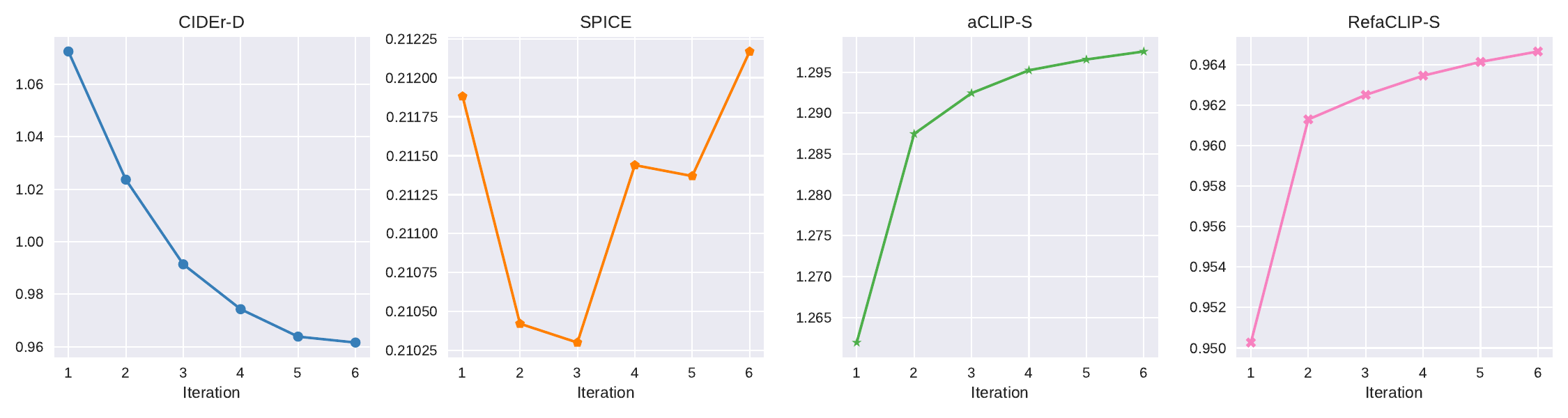}
    \vskip -1em
    \caption{Development of CIDEr-D, SPICE, aCLIP-S, and RefaCLIP-S for DAL on the MS-COCO validation set where we use RefaCLIP-S for quality filtering.}
    \label{fig:dal_metric_development}   
\end{figure}

\begin{figure}
    \centering
    \includegraphics[width=.8\textwidth]{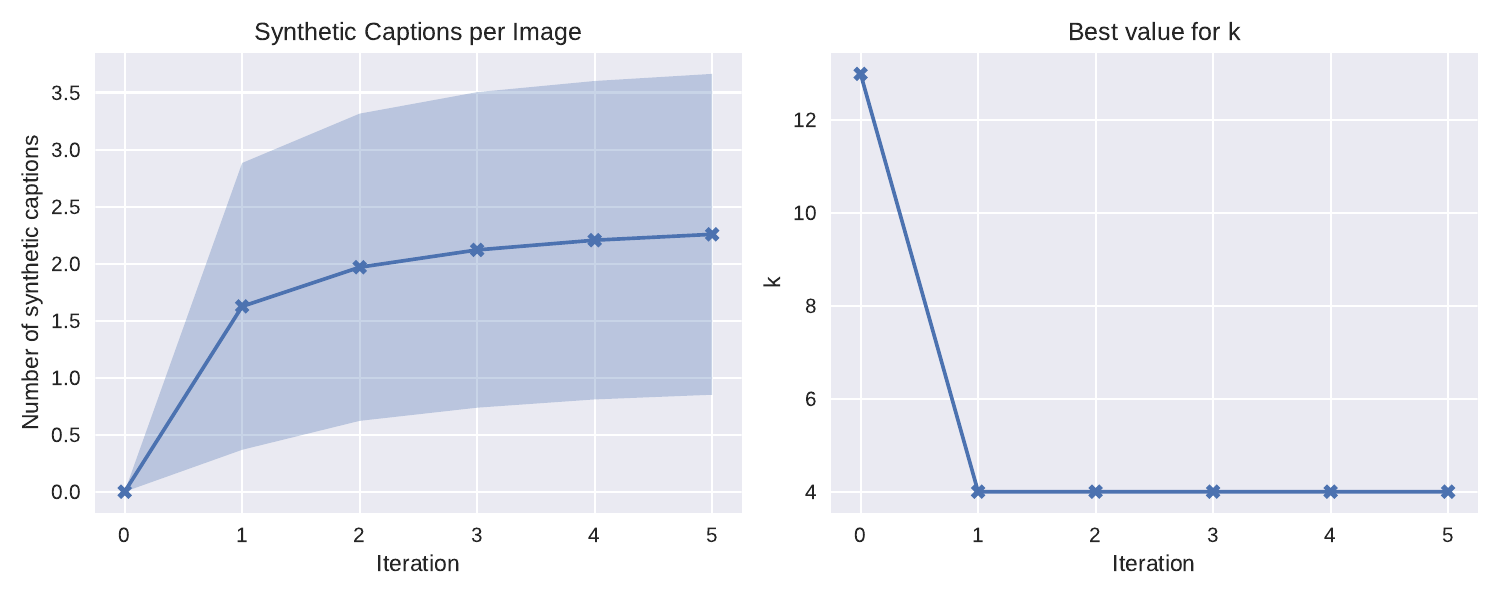}
    \vskip -1em
    \caption{Development of the hyperparameter $k$ and the number of synthetic captions per image during DAL on the MS-COCO dataset.}
    \label{fig:dal_synths_per_img}   
\end{figure}

We investigate the development of the different metrics after each iteration of DAL on the MS-COCO validation set in \cref{fig:dal_metric_development}. 
We observe that CIDEr-D constantly decreases, while SPICE fluctuates without changing significantly.
However, aCLIP-S and RefaCLIP-S exhibit a significant and monotonic improvement across every DAL iteration.
Further, we show the development of the hyperparameter $k$ during DAL and the number of synthetic captions that are on average provided to the LM for a given image in \cref{fig:dal_synths_per_img}.
We find that as soon as we add synthetic captions to the datastore (\cref{fig:dal_synths_per_img}, right), the best choice for $k$ on the validation set decreases from $k=13$ to $k=4$ and stagnates.
We hypothesize this is due to the increasing amount of synthetic captions that would otherwise be present in the prompt which might harm performance.
The number of synthetic captions in the prompt (\cref{fig:dal_synths_per_img}, left) generally increases with more iterations of DAL since more synthetic captions are added to the datastore.
Approximately two out of four captions in the prompt of the LM are synthetic, which amounts to 50\% of the captions in the prompt.
This number is similar across all iterations of DAL.
This means that the prompt to the LM is a balanced mix of human annotated captions and synthetically generated captions.
We believe that this is the desired behavior to ensure the generated captions do not diverge too much from ground truth references.
Note that this behavior naturally emerges during training and we did not control for this.

\begin{figure*}[t]
    \centering
    \includegraphics[width=\textwidth]{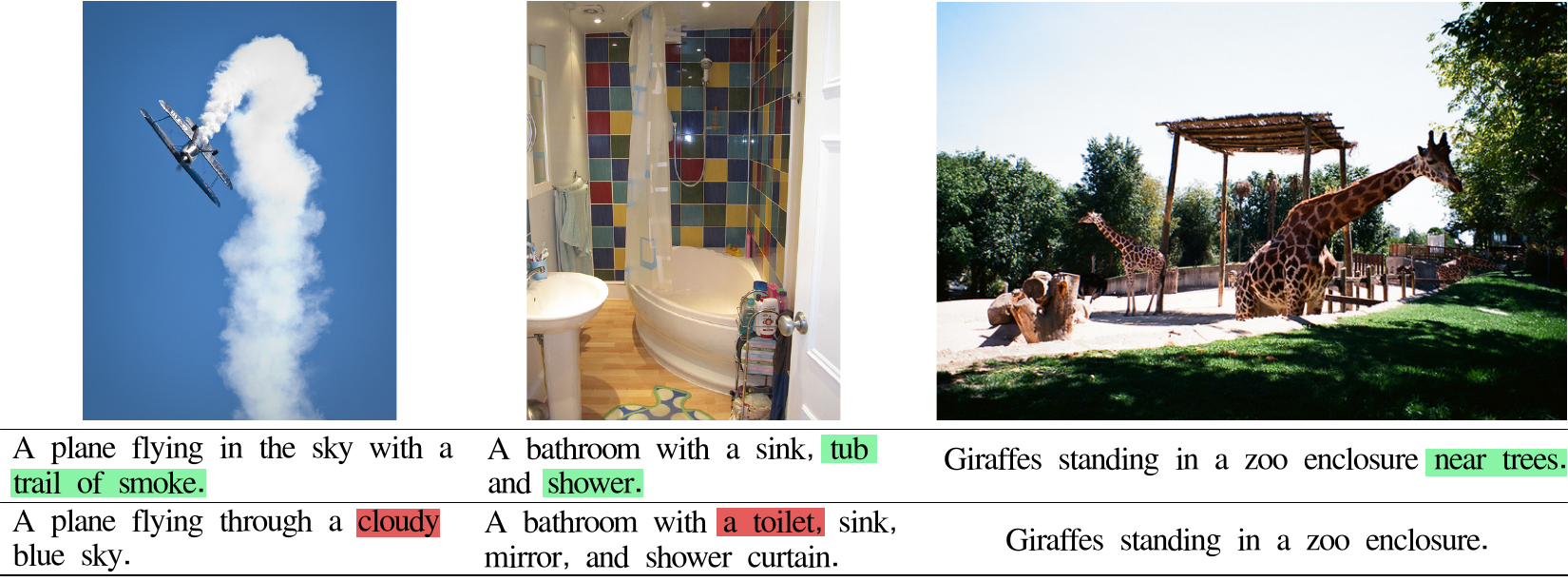}
    \vskip -.5em
    \caption{Captions generated via \OurMethod (bottom) and \DAL (top). Images were taken from the MS-COCO validation set.}
    \label{fig:qualitative_dal_captions}
\end{figure*}

Finally, we show some sample images from the MS-COCO test split and captions generated by \OurMethod and \DAL in \cref{fig:qualitative_dal_captions}.
We observe that \DAL generates more detailed captions, such as recognizing trees in \cref{fig:qualitative_dal_captions}, right.
Further, in some cases \DAL removes some imaginary content from captions, as showcased in \cref{fig:qualitative_dal_captions} left and middle.
We provide further examples in \cref{fig:qualitative_dal_captions_appendix}.

\paragraph{Image-captioning Metric}

We report extended results for caption evaluation and show additional results on the THumB dataset \citep{kasai_transparent_2022}.
THumB is a subset of MS-COCO images from the test split of \citet{karpathy_deep_2017} that contains human rankings for candidate captions.
Again, we compare our metrics against the current state-of-the-art metrics, namely CLIP+DN \citep{zhou_distribution_2023} and CLIP-score variants \citep[CLIP-S,RefCLIP-S]{hessel_clipscore_2021}.
We also include an ablation of CLIP+DN, called CLIP+DN* from \citet{zhou_distribution_2023} and an ablation for our metrics where we use the ViT-B/32 encoder \citep{dosovitskiy_image_2021}.
There are no published results for MID on THumB and SoftSPICE on Flickr8k-CF and THumB.
We observe a significant improvement of aCLIP-S and RefaCLIP-S over CLIP-S and RefCLIP-S.
However, CLIP+DN variants reach higher correlation with human judgements on THumB.
Interestingly, we find that the RN50$\times$64 based encoder generally correlates more strongly with human judgement than the ViT-B/32 encoder in both the reference-based, and the reference-free case.
These results suggest, that the best metric for evaluation depends on the dataset to evaluate on, as our reference-free metric outperformed CLIP+DN variants on the Flickr8k-Expert and Flickr8k-Crowdflower datasets.

\begin{table}[t]
\caption{Correlation with human judgement for different CLIP vision encoders measured via Kendall's $\tau_c$ for Flickr8k-E and $\tau_b$ for Flickr8k-CF both scaled by 100. The variance for the $\tau$ estimator only depends on sample size and is $3\text{e-}5$ for Flickr8k-E and $1\text{e-}5$ for Flickr8k-CF.}
\label{tab:metrics_encoder_ablation}
\vskip 0.15in
\begin{center}
\begin{small}
\begin{sc}
    \begin{tabular}{l| c c c |c}
    \toprule
    Method & Flickr8k-e & Flickr8k-cf & Thumb & Avg\\
    \midrule
    \multicolumn{5}{c}{Reference-free} \\
    \midrule
    $\text{aCLIP-S}_{\text{RN50}}$ & 54.4 & 34.9 & 18.6 & 36.0\\
    $\text{aCLIP-S}_{\text{RN101}}$ & 55.0 & 35.0 & 21.0 & 37.0 \\
    $\text{aCLIP-S}_{\text{RN50x4}}$ & 55.2 & 35.2 & 21.7 & 37.4 \\
    $\text{aCLIP-S}_{\text{RN50x16}}$ & 55.2 & 35.6 & 22.2 & 37.7 \\
    $\text{aCLIP}_{\text{RN50x64}}$ & 55.1 & 36.2 & 22.5 & 37.9 \\
    $\text{aCLIP}_{\text{ViT-B/32}}$ & 54.9 & 34.9 & 20.5 & 36.8 \\
    $\text{aCLIP-S}_{\text{ViT-B/16}}$ & 55.4 & 35.5 & 21.9 & 37.6 \\
    $\text{aCLIP-S}_{\text{ViT-L/14}}$ & \textbf{55.7} & 35.8 & 24.0 & 38.5 \\
    $\text{aCLIP-S}_{\text{ViT-L/14@336}}$ & 55.6 & \textbf{36.0} & \textbf{24.9} & \textbf{38.8} \\
    
    \midrule
    \multicolumn{5}{c}{Reference-based} \\
    \midrule
    $\text{RefaCLIP-S}_{\text{RN50}}$ & 54.8 & 35.5 & 20.4 & 36.9 \\
    $\text{RefaCLIP-S}_{\text{RN101}}$ & 55.4 & 35.5 & 22.7 & 37.9 \\
    $\text{RefaCLIP-S}_{\text{RN50x4}}$ & 55.5 & 35.8 & 23.4 & 38.2 \\
    $\text{RefaCLIP-S}_{\text{RN50x16}}$ & 55.6 & 36.0 & 23.5 & 38.4 \\
    $\text{RefaCLIP-S}_{\text{RN50x64}}$ & 55.5 & \textbf{36.7} & 24.3 & 38.8 \\
    $\text{RefaCLIP-S}_{\text{ViT-B/32}}$ & 55.3 & 35.4 & 21.7 & 37.5\\
    $\text{RefaCLIP-S}_{\text{ViT-B/16}}$ & 55.7 & 35.9 & 23.0 & 38.2 \\
    $\text{RefaCLIP-S}_{\text{ViT-L/14}}$ & \textbf{56.1} & 36.3 & 24.9 & 39.1 \\
    $\text{RefaCLIP-S}_{\text{ViT-L/14@336}}$ & 56.0 & 36.5 & \textbf{25.6} & \textbf{39.4} \\
    \bottomrule
\end{tabular}
\end{sc}
\end{small}
\end{center}
\end{table}

\section{Additional Qualitative Analysis}
\label{app:qualitative_analysis}

\begin{figure*}[h]
    \centering
    \includegraphics[width=.9\textwidth]{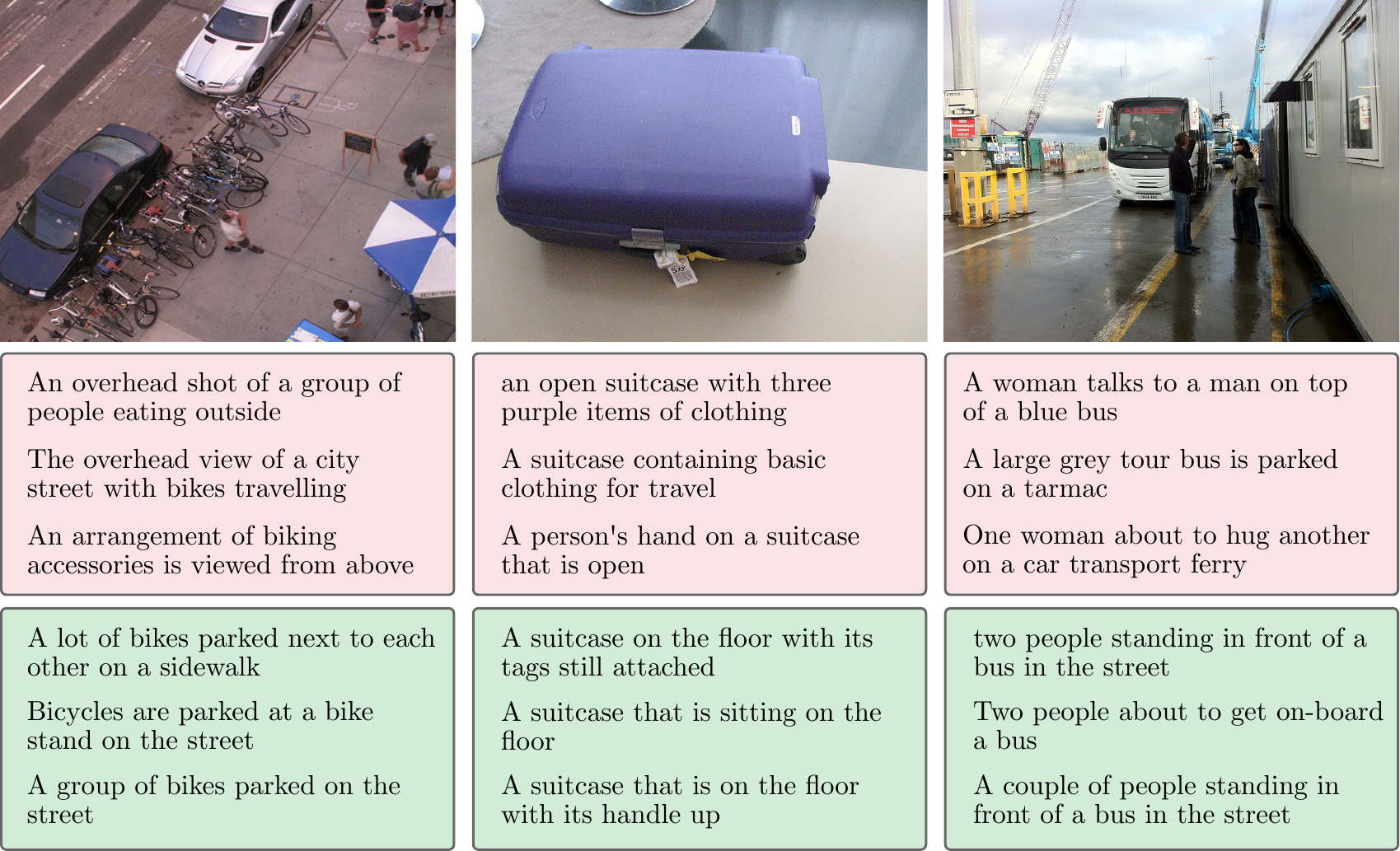}
    \vskip -0.5em
    \caption{Sample images and retrieved captions with (bottom) and without (top) our linear alignment to MS-COCO training data. We show three of the closest captions to an image. Images are taken from the MS-COCO validation set.}
    \label{fig:retrieval_example}
\end{figure*}

We show some examples for retrieval with and without our linear alignment in \cref{fig:retrieval_example}.
The top row shows the top-k samples for using off-the-shelf CLIP for retrieval, while the bottom row shows retrieval for our aligned CLIP.
After the linear alignment, the retrievals fit better to the image.
For example, CLIP assigns a high similarity to ``\emph{open suitcase}'' for the figure in the middle, although the suitcase in the image is closed.
Our aligned CLIP does not assign a high similarity to the same caption anymore, and retrieves more appropriate captions.

\begin{figure*}[t]
    \centering
    \includegraphics[width=\textwidth]{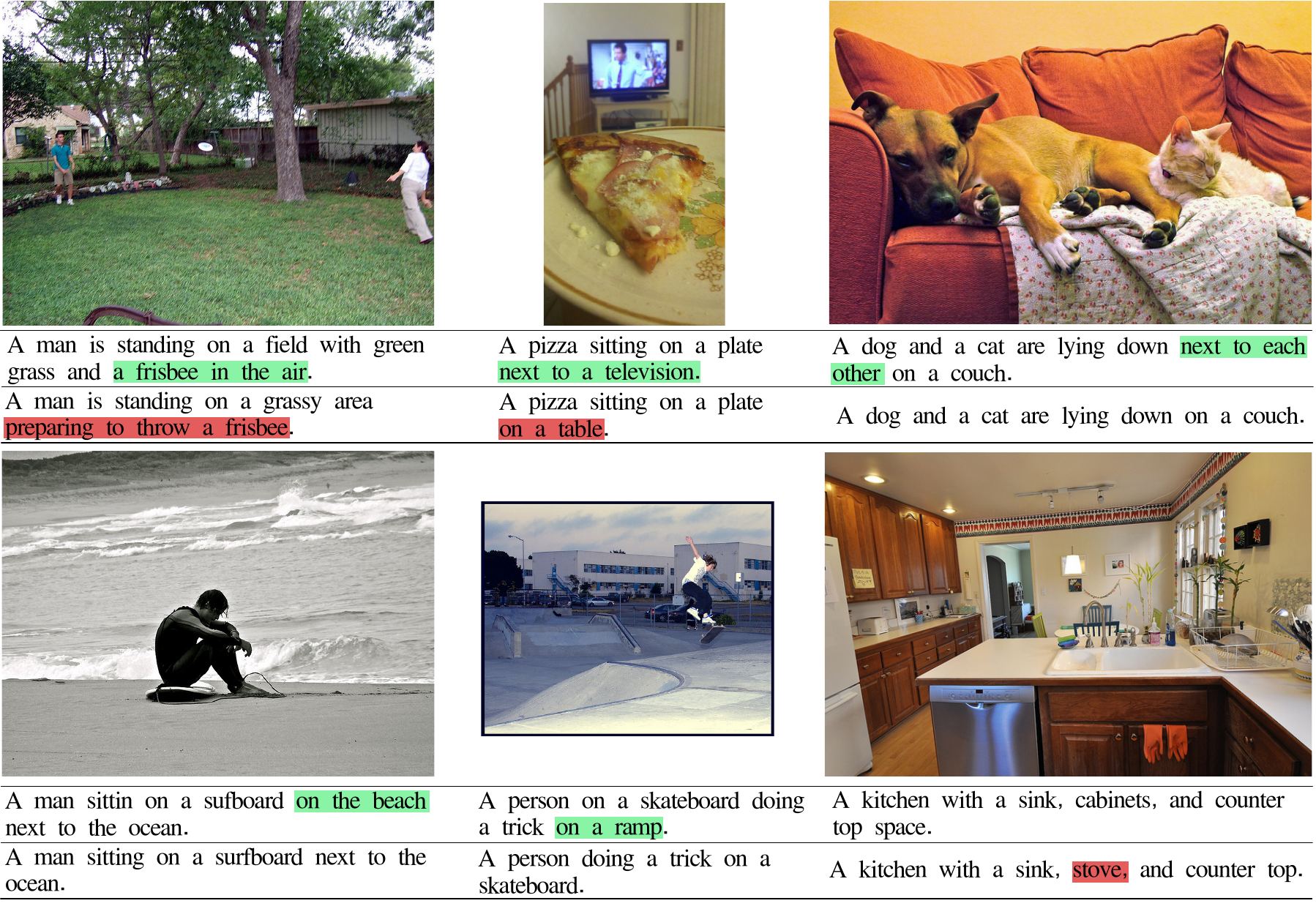}
    \vskip -1em
    \caption{Captions generated via \OurMethod (bottom) and \DAL (top). Images were taken from the MS-COCO validation set.}
    \label{fig:qualitative_dal_captions_appendix}
\end{figure*}

We show additional examples for captions generated after our DAL in \cref{fig:qualitative_dal_captions_appendix}.

\section{Hyperparameter Search}
\label{app:hyperparams}

\paragraph{Effect of different vision encoders}

We investigate the effect of different vision encoders on the captioning performance of \OurMethod on the MS-COCO validation set.
In this regard, we compare all publicly available encoder variants of CLIP, which comprise ViT-based \citep{dosovitskiy_image_2021}, as well as resnet-based \citep{he_deep_2016} architectures.
The best performing model for our retrieval-based image captioning is RN50$\times$64 (see \cref{tab:encoder_ablation}).
This corroborates our results for cross-modal retrieval, where RN50$\times$64 outperformed all other encoders \cref{sec:additional_results}.

\begin{table}[p]
\caption{Search over all publicly available CLIP vision encoder backbones evaluated on the MS-COCO validation set. We report mean and standard error for all settings. $|\theta|$ denotes the number of trainable parameters.}
\label{tab:encoder_ablation}
\vskip 0.15in
\begin{center}
\begin{small}
\begin{sc}
    \begin{tabular}{l c c c c c c c c}
    \toprule
    Vision Encoder & BLEU@1 & BLEU@4 & ROUGE-L & CIDEr-D & SPICE & $|\theta|$\\
    \midrule
    RN50 & 75.5 $\pm$ 0.2 & 28.0 $\pm$ 0.3 & 56.1 $\pm$ 0.2 & 97.0 $\pm$ 0.9 & 19.7 $\pm$ 0.1 & 1\,M\\
    RN101 & 74.6 $\pm$ 0.2 & 27.7 $\pm$ 0.3 & 56.1 $\pm$ 0.2 & 96.3 $\pm$ 0.9 & 19.4 $\pm$ 0.1 & 262\,K\\
    RN50x4 & 75.4 $\pm$ 0.2 & 28.5 $\pm$ 0.3 & 56.6 $\pm$ 0.2 & 99.2 $\pm$ 0.9  & 19.9 $\pm$ 0.1 & 410\,K\\
    RN50x16 & 76.4 $\pm$ 0.2 & 29.3 $\pm$ 0.4 & 57.0 $\pm$ 0.2 & 102.5 $\pm$ 0.9 & 20.4 $\pm$ 0.1 & 590\,K\\
    RN50x64 & 77.7 $\pm$ 0.2 & 30.5 $\pm$ 0.4 & 58.0 $\pm$ 0.2 & 107.3 $\pm$ 1.0 & 21.2 $\pm$ 0.1 & 1\,M\\
    ViT-B/32 & 75.2 $\pm$ 0.2 & 27.9 $\pm$ 0.3 & 56.0 $\pm$ 0.2 & 96.4 $\pm$ 0.9 & 19.4 $\pm$ 0.1 & 262\,K\\
    ViT-B/16 & 76.2 $\pm$ 0.2 & 29.0 $\pm$ 0.3 & 56.7 $\pm$ 0.2 & 101.2 $\pm$ 0.9 & 20.0 $\pm$ 0.1 & 262\,K\\
    ViT-L/14 & 77.0 $\pm$ 0.2 & 29.9 $\pm$ 0.4 & 57.4 $\pm$ 0.2 & 104.7 $\pm$ 1.0 & 20.6 $\pm$ 0.1 & 590\,K\\
    ViT-L/14@336px & 77.4 $\pm$ 0.2 & 30.3 $\pm$ 0.4 & 57.7 $\pm$ 0.2 & 105.8 $\pm$ 0.9 & 20.8 $\pm$ 0.1 & 590\,K\\
    \bottomrule
    \end{tabular}
\end{sc}
\end{small}
\end{center}
\vskip -0.1in
\end{table}

\paragraph{Top-k retrieval}

We search over different values for our hyperparameters $k$ on the MS-COCO, Flickr30k, VizWiz, and MSRVTT validation sets.
We report results in \cref{tab:coco_hyperparams} and \cref{tab:flickr_hyperparams} for MS-COCO, and Flickr30k, respectively.
The results for VizWiz and MSRVTT are shown in \cref{tab:vizwiz_k_search}, and \cref{tab:msrvtt_k_search}, respectively.
For searching over values for $k$ we use greedy decoding, to isolate the effect of the hyperparameter.

\begin{table}[p]
\caption{Hyperparameter Search for $k$ on the MS-COCO validation set for different levels of language abstraction using our semantic mapping computed via OLS. We report mean and standard error for all settings. We select the best k according to CIDEr-D score.}
\label{tab:coco_hyperparams}
\vskip 0.15in
\begin{center}
\begin{small}
\begin{sc}
    \begin{tabular}{c c c c c c c c}
    \toprule
    $k$ & BLEU@1 & BLEU@4 & ROUGE-L & CIDEr-D & SPICE\\
    \midrule
    \multicolumn{6}{c}{Single Captions}\\
    \midrule
    10 & 77.4 $\pm$ 0.2 & 30.4 $\pm$ 0.4 & 57.6 $\pm$ 0.2 & 105.2 $\pm$ 1.0 & 20.9 $\pm$ 0.1\\
    11 & 77.4 $\pm$ 0.2 & 30.4 $\pm$ 0.4 & 57.7 $\pm$ 0.2 & 105.4 $\pm$ 1.0 & 20.9 $\pm$ 0.1\\
    12 & 77.4 $\pm$ 0.2 & 30.3 $\pm$ 0.4 & 57.7 $\pm$ 0.2 & 105.2 $\pm$ 1.0 & 20.9 $\pm$ 0.1\\
    13 & 77.4 $\pm$ 0.2 & 30.5 $\pm$ 0.4 & 57.7 $\pm$ 0.2 & 105.5 $\pm$ 1.0 & 20.8 $\pm$ 0.1\\
    14 & 77.4 $\pm$ 0.2 & 30.5 $\pm$ 0.4 & 57.8 $\pm$ 0.2 & 105.4 $\pm$ 1.0 & 20.8 $\pm$ 0.1\\
    15 & 77.3 $\pm$ 0.2 & 30.5 $\pm$ 0.4 & 57.7 $\pm$ 0.2 & 105.4 $\pm$ 1.0 & 20.9 $\pm$ 0.1\\
    16 & 77.2 $\pm$ 0.2 & 30.4 $\pm$ 0.4 & 57.7 $\pm$ 0.2 & 105.4 $\pm$ 1.0 & 20.8 $\pm$ 0.1\\
    17 & 77.2 $\pm$ 0.2 & 30.2 $\pm$ 0.4 & 57.6 $\pm$ 0.2 & 104.9 $\pm$ 1.0 & 20.9 $\pm$ 0.1\\
    \midrule
    \multicolumn{6}{c}{All Captions}\\
    \midrule
    1 & 72.7 $\pm$ 0.2 & 24.8 $\pm$ 0.3 & 53.9 $\pm$ 0.2 & 87.0 $\pm$ 0.9 & 18.0 $\pm$ 0.1 \\
    2 & 73.7 $\pm$ 0.2 & 26.4 $\pm$ 0.3 & 54.7 $\pm$ 0.2 & 90.8 $\pm$ 0.9 & 18.2 $\pm$ 0.1 \\
    3 & 74.0 $\pm$ 0.2 & 26.4 $\pm$ 0.3 & 54.8 $\pm$ 0.2 & 91.0 $\pm$ 0.9 & 18.2 $\pm$ 0.1 \\
    4 & 74.0 $\pm$ 0.2 & 26.6 $\pm$ 0.3 & 55.0 $\pm$ 0.2 & 91.3 $\pm$ 0.9 & 18.5 $\pm$ 0.1 \\
    5 & 74.0 $\pm$ 0.2 & 26.9 $\pm$ 0.3 & 55.1 $\pm$ 0.2 & 91.6 $\pm$ 0.9 & 18.4 $\pm$ 0.1 \\
    \midrule
    \multicolumn{6}{c}{Localized Narratives}\\
    \midrule
    1 & 55.3 $\pm$ 0.3 & 11.7 $\pm$ 0.2 & 43.1 $\pm$ 0.2 & 45.4 $\pm$ 0.6 & 11.9 $\pm$ 0.1 \\
    2 & 54.3 $\pm$ 0.3 & 11.8 $\pm$ 0.2 & 43.0 $\pm$ 0.2 & 48.0 $\pm$ 0.7 & 13.2 $\pm$ 0.1 \\
    3 & 53.8 $\pm$ 0.3 & 12.3 $\pm$ 0.2 & 43.0 $\pm$ 0.2 & 50.9 $\pm$ 0.7 & 14.0 $\pm$ 0.1 \\
    4 & 53.0 $\pm$ 0.3 & 12.1 $\pm$ 0.2 & 42.7 $\pm$ 0.2 & 51.7 $\pm$ 0.7 & 14.3 $\pm$ 0.1 \\
    5 & 52.5 $\pm$ 0.3 & 12.0 $\pm$ 0.2 & 42.6 $\pm$ 0.2 & 52.6 $\pm$ 0.7 & 14.4 $\pm$ 0.1 \\
    6 & 52.0 $\pm$ 0.3 & 12.3 $\pm$ 0.2 & 42.6 $\pm$ 0.2 & 53.1 $\pm$ 0.7 & 14.6 $\pm$ 0.1 \\
    \bottomrule
    \end{tabular}
\end{sc}
\end{small}
\end{center}
\vskip -0.1in
\end{table}

\begin{table}[p]
\caption{Hyperparameter Search for $k$ on the Flickr30k validation set for different levels of language abstraction using our semantic mapping computed via OLS. We report mean and standard error for all settings.}
\label{tab:flickr_hyperparams}
\vskip 0.15in
\begin{center}
\begin{small}
\begin{sc}
    \begin{tabular}{l c c c c c c c}
    \toprule
    $k$ & BLEU@1 & BLEU@4 & ROUGE-L & CIDEr-D & SPICE \\
    \midrule
    \multicolumn{6}{c}{Single Captions}\\
    \midrule
    10 & 74.8 $\pm$ 0.5 & 26.4 $\pm$ 0.7 & 54.5 $\pm$ 0.4 & 63.6 $\pm$ 1.9 & 15.5 $\pm$ 0.3 \\
    11 & 74.7 $\pm$ 0.5 & 26.3 $\pm$ 0.7 & 54.5 $\pm$ 0.4 & 64.4 $\pm$ 2.0 & 15.6 $\pm$ 0.3 \\
    12 & 74.4 $\pm$ 0.5 & 26.2 $\pm$ 0.7 & 54.6 $\pm$ 0.4 & 64.6 $\pm$ 1.9 & 15.5 $\pm$ 0.3 \\
    13 & 74.2 $\pm$ 0.5 & 26.1 $\pm$ 0.7 & 54.6 $\pm$ 0.4 & 64.4 $\pm$ 1.9 & 15.5 $\pm$ 0.3 \\
    14 & 74.6 $\pm$ 0.5 & 26.2 $\pm$ 0.7 & 54.3 $\pm$ 0.4 & 64.4 $\pm$ 1.9 & 15.6 $\pm$ 0.3 \\
    15 & 74.3 $\pm$ 0.5 & 26.3 $\pm$ 0.7 & 54.5 $\pm$ 0.4 & 64.8 $\pm$ 1.9 & 15.6 $\pm$ 0.3 \\
    16 & 75.0 $\pm$ 0.5 & 26.7 $\pm$ 0.7 & 54.7 $\pm$ 0.4 & 64.6 $\pm$ 1.9 & 15.8 $\pm$ 0.3 \\
    17 & 74.5 $\pm$ 0.5 & 26.9 $\pm$ 0.7 & 54.8 $\pm$ 0.4 & 65.5 $\pm$ 1.9 & 15.6 $\pm$ 0.3 \\
    18 & 74.9 $\pm$ 0.5 & 26.8 $\pm$ 0.7 & 54.8 $\pm$ 0.4 & 66.2 $\pm$ 2.0 & 15.7 $\pm$ 0.3 \\
    19 & 74.4 $\pm$ 0.5 & 26.9 $\pm$ 0.7 & 54.8 $\pm$ 0.4 & 65.6 $\pm$ 1.9 & 15.8 $\pm$ 0.3 \\
    \midrule
    \multicolumn{6}{c}{All Captions}\\
    \midrule
    1 & 65.8 $\pm$ 0.5 & 20.3 $\pm$ 0.7 & 49.8 $\pm$ 0.4 & 48.7 $\pm$ 1.8 & 13.4 $\pm$ 0.3 \\
    2 & 67.9 $\pm$ 0.5 & 21.5 $\pm$ 0.7 & 50.5 $\pm$ 0.5 & 52.2 $\pm$ 1.8 & 13.9 $\pm$ 0.3 \\
    3 & 68.1 $\pm$ 0.5 & 22.0 $\pm$ 0.7 & 51.0 $\pm$ 0.4 & 53.2 $\pm$ 1.9 & 13.7 $\pm$ 0.3 \\
    4 & 69.6 $\pm$ 0.5 & 23.0 $\pm$ 0.7 & 51.4 $\pm$ 0.4 & 54.4 $\pm$ 1.9 & 14.1 $\pm$ 0.3 \\
    5 & 69.0 $\pm$ 0.5 & 23.0 $\pm$ 0.7 & 51.3 $\pm$ 0.4 & 54.5 $\pm$ 1.9 & 14.2 $\pm$ 0.3 \\
    \midrule
    \multicolumn{6}{c}{Localized Narratives}\\
    \midrule
    1 & 54.2 $\pm$ 0.6 & 9.0 $\pm$ 0.4 & 40.4 $\pm$ 0.4 & 24.4 $\pm$ 1.3 & 8.1 $\pm$ 0.2 \\
    2 & 52.6 $\pm$ 0.6 & 8.6 $\pm$ 0.4 & 39.3 $\pm$ 0.4 & 23.3 $\pm$ 1.1 & 8.4 $\pm$ 0.2 \\
    3 & 52.5 $\pm$ 0.6 & 9.5 $\pm$ 0.4 & 39.6 $\pm$ 0.4 & 25.4 $\pm$ 1.2 & 8.9 $\pm$ 0.2 \\
    4 & 51.7 $\pm$ 0.6 & 9.6 $\pm$ 0.4 & 39.3 $\pm$ 0.4 & 26.0 $\pm$ 1.2 & 9.1 $\pm$ 0.2 \\
    5 & 51.9 $\pm$ 0.6 & 9.6 $\pm$ 0.4 & 39.1 $\pm$ 0.4 & 25.6 $\pm$ 1.2 & 9.0 $\pm$ 0.2 \\
    \bottomrule
    \end{tabular}
\end{sc}
\end{small}
\end{center}
\vskip -0.1in
\end{table}

\begin{table}[p]
\caption{Hyperparameter Search for $k$ on the VizWiz validation set for ReCap with our linear alignment. We report mean and standard error for all settings. We select the best k according to CIDEr-D score.}
\label{tab:vizwiz_k_search}
\vskip 0.15in
\begin{center}
\begin{small}
\begin{sc}
    \begin{tabular}{c c c c c c c c}
    \toprule
    $k$ & BLEU@1 & BLEU@4 & ROUGE-L & CIDEr-D & SPICE\\
    \midrule
    1 & 61.8 $\pm$ 0.2 & 15.5 $\pm$ 0.2 & 43.1 $\pm$ 0.2 & 48.5 $\pm$ 0.6 & 12.1 $\pm$ 0.1\\
    2 & 61.8 $\pm$ 0.2 & 16.5 $\pm$ 0.2 & 44.8 $\pm$ 0.2 & 50.9 $\pm$ 0.7 & 13.1 $\pm$ 0.1\\
    3 & 62.5 $\pm$ 0.2 & 16.9 $\pm$ 0.2 & 45.3 $\pm$ 0.2 & 51.1 $\pm$ 0.7 & 13.0 $\pm$ 0.1\\
    4 & 63.2 $\pm$ 0.2 & 17.5 $\pm$ 0.2 & 45.8 $\pm$ 0.2 & 52.7 $\pm$ 0.7 & 13.0 $\pm$ 0.1\\
    5 & 63.3 $\pm$ 0.2 & 17.5 $\pm$ 0.2 & 45.8 $\pm$ 0.2 & 52.6 $\pm$ 0.7 & 13.1 $\pm$ 0.1\\
    6 & 63.3 $\pm$ 0.2 & 17.6 $\pm$ 0.2 & 45.9 $\pm$ 0.2 & 52.4 $\pm$ 0.7 & 13.0 $\pm$ 0.1\\
    7 & 63.0 $\pm$ 0.2 & 17.5 $\pm$ 0.2 & 45.8 $\pm$ 0.2 & 51.7 $\pm$ 0.7 & 12.9 $\pm$ 0.1\\
    8 & 62.8 $\pm$ 0.2 & 17.5 $\pm$ 0.2 & 45.8 $\pm$ 0.2 & 51.6 $\pm$ 0.7 & 12.8 $\pm$ 0.1\\
    9 & 62.9 $\pm$ 0.2 & 17.5 $\pm$ 0.2 & 45.9 $\pm$ 0.2 & 51.3 $\pm$ 0.7 & 12.9 $\pm$ 0.1\\
    10 & 62.1 $\pm$ 0.2 & 17.0 $\pm$ 0.2 & 45.5 $\pm$ 0.2 & 50.3 $\pm$ 0.6 & 12.8 $\pm$ 0.1\\
    \bottomrule
    \end{tabular}
\end{sc}
\end{small}
\end{center}
\vskip -0.1in
\end{table}

\begin{table}[p]
\caption{Hyperparameter Search for $k$ on the MSRVTT validation set for ReCap with our linear alignment. We report mean and standard error for all settings. We select the best k according to CIDEr-D score.}
\label{tab:msrvtt_k_search}
\vskip 0.15in
\begin{center}
\begin{small}
\begin{sc}
    \begin{tabular}{c c c c c c c c}
    \toprule
    $k$ & BLEU@1 & BLEU@4 & ROUGE-L & CIDEr-D & SPICE\\
    \midrule
    3 & 26.9 $\pm$ 0.1 & 4.8 $\pm$ 0.0 & 25.7 $\pm$ 0.1 & 36.6 $\pm$ 0.4 & 14.2 $\pm$ 0.1 \\
    4 & 26.9 $\pm$ 0.1 & 4.8 $\pm$ 0.0 & 25.7 $\pm$ 0.1 & 36.6 $\pm$ 0.4 & 14.2 $\pm$ 0.1\\
    5 & 27.1 $\pm$ 0.1 & 4.9 $\pm$ 0.0 & 25.8 $\pm$ 0.1 & 36.7 $\pm$ 0.4 & 14.1 $\pm$ 0.1\\
    6 & 27.1 $\pm$ 0.1 & 4.9 $\pm$ 0.0 & 25.8 $\pm$ 0.1 & 36.4 $\pm$ 0.4 & 14.0 $\pm$ 0.1\\
    7 & 27.0 $\pm$ 0.1 & 4.9 $\pm$ 0.0 & 25.9 $\pm$ 0.1 & 36.4 $\pm$ 0.3 & 13.9 $\pm$ 0.1\\
    8 & 27.0 $\pm$ 0.1 & 4.9 $\pm$ 0.0 & 25.9 $\pm$ 0.1 & 36.7 $\pm$ 0.4 & 13.8 $\pm$ 0.1\\
    \bottomrule
    \end{tabular}
\end{sc}
\end{small}
\end{center}
\vskip -0.1in
\end{table}

\begin{table}[h]
\caption{Hyperparameter Search for $k$ on the chest-xrays validation set for ReCap with our linear alignment. We report mean and standard error for all settings. We select the best k according to CIDEr-D score.}
\label{tab:chest_xrays_k_search}
\vskip 0.15in
\begin{center}
\begin{small}
\begin{sc}
    \begin{tabular}{c c c c c c c c}
    \toprule
    $k$ & BLEU@1 & BLEU@4 & ROUGE-L & CIDEr-D & SPICE\\
    \midrule
    1 & 35.4 $\pm$ 0.8 & 7.8 $\pm$ 0.4 & 28.7 $\pm$ 0.7 & 14.5 $\pm$ 1.1 & 8.6 $\pm$ 0.4 \\
    2 & 34.8 $\pm$ 0.8 & 7.8 $\pm$ 0.4 & 29.9 $\pm$ 0.7 & 14.9 $\pm$ 1.1 & 9.2 $\pm$ 0.4 \\
    3 & 34.1 $\pm$ 0.8 & 7.9 $\pm$ 0.4 & 30.0 $\pm$ 0.7 & 14.7 $\pm$ 1.1 & 9.4 $\pm$ 0.4 \\
    4 & 32.2 $\pm$ 0.8 & 7.3 $\pm$ 0.4 & 30.3 $\pm$ 0.7 & 15.6 $\pm$ 1.2 & 10.0 $\pm$ 0.4 \\
    5 & 30.7 $\pm$ 0.8 & 6.8 $\pm$ 0.4 & 30.9 $\pm$ 0.7 & 16.4 $\pm$ 1.4 & 10.3 $\pm$ 0.4 \\
    6 & 29.2 $\pm$ 0.8 & 6.6 $\pm$ 0.4 & 30.4 $\pm$ 0.7 & 16.0 $\pm$ 1.3 & 10.2 $\pm$ 0.4 \\
    7 & 27.1 $\pm$ 0.8 & 5.8 $\pm$ 0.4 & 29.4 $\pm$ 0.6 & 12.8 $\pm$ 1.0 & 9.7 $\pm$ 0.4 \\
    \bottomrule
    \end{tabular}
\end{sc}
\end{small}
\end{center}
\vskip -0.1in
\end{table}

\paragraph{Language-model scales}

We evaluate FLAN-T5 model sizes of 80\,M, 250\,M, 720\,M, 3\,B, and 11\,B scales.
Further, we include decoder-only LMs, such as GPT-2 \citep{radford_language_2018}, GPT-J \citep{wang_gpt-j_2021}, and Llama 7B \citep{touvron_llama_2023}.
The results can be observed in \cref{tab:lm_ablation}.
Our results show that there is not much performance gain going from FLAN-T5-LARGE to FLAN-T5-XXL.
We suspect this is due to the design of the prompt which apparently suits FLAN-T5-LARGE particularly well.
Surprisingly, even the small variant of FLAN-T5 reaches a CIDEr-D score above 90, which amounts to decent captioning quality.

Our results for decoder-only LMs show that they generally perform worse than encoder-decoder ones.
We found that decoder-only models are generally more sensitive to prompt ordering, which was also found in prior works \citep{zhao_calibrate_2021}.
Perhaps surprisingly, GPT-J outperforms the recently proposed Llama, which reaches performance on-par with GPT-2.
Generally, we belive that we could improve performance of larger models by more extensive prompt tuning.
However, remarkably, FLAN-T5 performs really well in our setup without the need for extensive prompt tuning.

\begin{table}[p]
\caption{Comparison of different language models on the MS-COCO validation set.  We report mean and standard error for all settings.}
\label{tab:lm_ablation}
\vskip 0.15in
\begin{center}
\begin{small}
\begin{sc}
    \begin{tabular}{l | c c c c c}
    \toprule
    Model & BLEU@1 & BLEU@4 & ROUGE-L & CIDEr-D & SPICE \\
    \midrule
    \multicolumn{6}{c}{Encoder-Decoder}\\
    \midrule
    FLAN-T5-SMALL & 63.9 $\pm$ 0.3 & 23.3 $\pm$ 0.3 & 55.0 $\pm$ 0.2 & 93.9 $\pm$ 1.0 & 20.5 $\pm$ 0.1\\
    FLAN-T5-BASE & 72.5 $\pm$ 0.2 & 27.1 $\pm$ 0.3 & 56.7 $\pm$ 0.2 & 100.0 $\pm$ 0.9 & 20.7 $\pm$ 0.1 \\
    FLAN-T5-LARGE & 77.7 $\pm$ 0.2 & 30.5 $\pm$ 0.4 & 58.0 $\pm$ 0.2 & 107.3 $\pm$ 1.0 & 21.2 $\pm$ 0.1 \\
    FLAN-T5-XL & 76.1 $\pm$ 0.2 & 29.4 $\pm$ 0.4 & 56.7 $\pm$ 0.2  & 104.7 $\pm$ 0.9 & 20.8 $\pm$ 0.1 \\
    FLAN-T5-XXL & 77.1 $\pm$ 0.2 & 30.2 $\pm$ 0.4 & 57.4 $\pm$ 0.2 & 107.0 $\pm$ 1.0  & 21.0 $\pm$ 0.1 \\
    \midrule
    \multicolumn{6}{c}{Decoder-only}\\
    \midrule
    GPT-2 & 64.9 $\pm$ 0.3 & 24.1 $\pm$ 0.3 & 49.5 $\pm$ 0.2 & 86.8 $\pm$ 0.9 & 19.1 $\pm$ 0.1 \\
    GPT-J 6B & 71.1 $\pm$ 0.3 & 29.1 $\pm$ 0.4 & 51.4 $\pm$ 0.2 & 97.5 $\pm$ 1.0 & 19.6 $\pm$ 0.1 \\
    Llama 7B & 61.5 $\pm$ 0.3 & 23.1 $\pm$ 0.3 & 49.3 $\pm$ 0.2 & 86.4 $\pm$ 0.9 & 19.5 $\pm$ 0.1 \\
    \bottomrule
    \end{tabular}
\end{sc}
\end{small}
\end{center}
\vskip -0.1in
\end{table}

\paragraph{Different decoding strategies}

As illustrated by \citep{holtzman_curious_2020}, the decoding strategy substantially affects human approval of generated captions.
Therefore, we evaluate different decoding strategies, including greedy decoding, sampling, top-k sampling, and nucleus sampling.
First, we search over different temperatures $\tau$ and number of generated captions $l$ for nucleus sampling \citep{holtzman_curious_2020}.
After sampling $l$ captions from the LM, we select the highest scoring one according to our aligned CLIP.
To find the best parameters $\tau$ and $l$ we set $k$ to the best value we found in the preceeding gridsearch with greedy decoding.
Results are reported in \cref{tab:temp_ablation_coco}, and \cref{tab:temp_ablation_flickr} for MS-COCO, and Flickr30k, respectively.
The results for VizWiz and MSRVTT are shown in \cref{tab:temp_search_vizwiz}, and \cref{tab:temp_search_msrvtt}, respectively.

\begin{table}[p]
\caption{Comparison of different values for temperature of nucleus sampling on the Flickr30k validation set for $k=18$}
\label{tab:temp_ablation_flickr}
\vskip 0.15in
\begin{center}
\begin{small}
\begin{sc}
    \begin{tabular}{c | c | c c c c c c c}
    \toprule
    Temperature & Samples & BLEU@1 & BLEU@4 & ROUGE-L & CIDEr-D & SPICE\\
    \midrule
    1.0 & 1 & 74.8 $\pm$ 0.5 & 26.8 $\pm$ 0.7 & 54.6 $\pm$ 0.4 & 65.0 $\pm$ 1.9 & 15.8 $\pm$ 0.3\\\midrule
    0.1 & 10 & 75.2 $\pm$ 0.5 & 27.5 $\pm$ 0.7 & 55.2 $\pm$ 0.4 & 68.7 $\pm$ 2.0 & 16.5 $\pm$ 0.3 \\
    0.3 & 10 & 74.5 $\pm$ 0.5 & 26.6 $\pm$ 0.7 & 55.2 $\pm$ 0.4 & 68.4 $\pm$ 1.9 & 16.8 $\pm$ 0.3 \\
    0.5 & 10 & 73.8 $\pm$ 0.5 & 25.6 $\pm$ 0.7 & 54.6 $\pm$ 0.4 & 68.4 $\pm$ 2.1 & 17.0 $\pm$ 0.3 \\
    \midrule
    0.1 & 20 & 75.3 $\pm$ 0.5 & 27.1 $\pm$ 0.7 & 55.2 $\pm$ 0.4 & 68.7 $\pm$ 1.9 & 16.5 $\pm$ 0.3 \\
    0.3 & 20 & 74.4 $\pm$ 0.5 & 26.6 $\pm$ 0.7 & 55.2 $\pm$ 0.4 & 69.3 $\pm$ 2.0 & 16.9 $\pm$ 0.3 \\
    0.5 & 20 & 73.4 $\pm$ 0.5 & 25.2 $\pm$ 0.7 & 54.6 $\pm$ 0.4 & 68.3 $\pm$ 2.0 & 17.3 $\pm$ 0.3 \\
    \midrule
    0.1 & 30 & 75.5 $\pm$ 0.5 & 27.5 $\pm$ 0.7 & 55.3 $\pm$ 0.4 & 68.7 $\pm$ 2.0 & 16.6 $\pm$ 0.3 \\
    0.3 & 30 & 74.2 $\pm$ 0.5 & 26.4 $\pm$ 0.7 & 55.4 $\pm$ 0.4 & 68.9 $\pm$ 2.0 & 17.2 $\pm$ 0.3 \\
    0.5 & 30 & 72.9 $\pm$ 0.5 & 24.4 $\pm$ 0.7 & 54.4 $\pm$ 0.4 & 67.7 $\pm$ 2.0 & 17.3 $\pm$ 0.3 \\
    \bottomrule
    \end{tabular}
\end{sc}
\end{small}
\end{center}
\vskip -0.1in
\end{table}

\begin{table}[p]
\caption{Comparison of different values for temperature of nucleus sampling on the MS-COCO validation set for $k=13$.}
\label{tab:temp_ablation_coco}
\vskip 0.15in
\begin{center}
\begin{small}
\begin{sc}
    \begin{tabular}{c | c | c c c c c c c}
    \toprule
    Temperature & Samples & BLEU@1 & BLEU@4 & ROUGE-L & CIDEr-D & SPICE\\
    \midrule
    0.0 & n/a & 77.4 $\pm$ 0.2 & 30.5 $\pm$ 0.4 & 57.7 $\pm$ 0.2 & 105.5 $\pm$ 1.0 & 20.8 $\pm$ 0.1\\\midrule
    0.1 & 10 & 77.7 $\pm$ 0.2 & 30.5 $\pm$ 0.4 & 58.0 $\pm$ 0.2 & 107.3 $\pm$ 1.0 & 21.2 $\pm$ 0.1 \\
    0.3 & 10 & 77.3 $\pm$ 0.2 & 29.9 $\pm$ 0.4 & 57.9 $\pm$ 0.2 & 106.8 $\pm$ 0.9 & 21.4 $\pm$ 0.1 \\
    0.5 & 10 & 76.5 $\pm$ 0.2 & 29.0 $\pm$ 0.3 & 57.3 $\pm$ 0.2 & 104.5 $\pm$ 0.9 & 21.3 $\pm$ 0.1 \\
    \midrule
    0.1 & 20 & 77.6 $\pm$ 0.2 & 30.4 $\pm$ 0.4 & 57.9 $\pm$ 0.2 & 107.2 $\pm$ 1.0 & 21.2 $\pm$ 0.1 \\
    0.3 & 20 & 77.2 $\pm$ 0.2 & 29.7 $\pm$ 0.3 & 57.8 $\pm$ 0.2 & 106.2 $\pm$ 0.9 & 21.4 $\pm$ 0.1 \\
    0.5 & 20 & 76.4 $\pm$ 0.2 & 28.6 $\pm$ 0.3 & 57.1 $\pm$ 0.2 & 103.9 $\pm$ 0.9 & 21.4 $\pm$ 0.1 \\
    \midrule
    0.1 & 30 & 77.6 $\pm$ 0.2 & 30.4 $\pm$ 0.4 & 57.9 $\pm$ 0.2 & 107.1 $\pm$ 0.9 & 21.2 $\pm$ 0.1 \\
    0.3 & 30 & 77.1 $\pm$ 0.2 & 29.5 $\pm$ 0.3 & 57.7 $\pm$ 0.2 & 106.1 $\pm$ 0.9 & 21.4 $\pm$ 0.1 \\
    0.5 & 30 & 76.4 $\pm$ 0.2 & 28.3 $\pm$ 0.3 & 57.1 $\pm$ 0.2 & 103.3 $\pm$ 0.9 & 21.6 $\pm$ 0.1 \\
    \bottomrule
    \end{tabular}
\end{sc}
\end{small}
\end{center}
\vskip -0.1in
\end{table}

\begin{table}[]
\caption{Comparison of different values for temperature of nucleus sampling on the VizWiz validation set for $k=4$.}
\label{tab:temp_search_vizwiz}
\vskip 0.15in
\begin{center}
\begin{small}
\begin{sc}
    \begin{tabular}{c | c | c c c c c c c}
    \toprule
    Temperature & Samples & BLEU@1 & BLEU@4 & ROUGE-L & CIDEr-D & SPICE\\
    \midrule
    0.0 & n/a & 63.2 $\pm$ 0.2 & 17.5 $\pm$ 0.2 & 45.8 $\pm$ 0.2 & 52.7 $\pm$ 0.7 & 13.0 $\pm$ 0.1\\\midrule
    0.1 & 10 & 64.5 $\pm$ 0.2 & 17.9 $\pm$ 0.2 & 46.3 $\pm$ 0.2 & 54.7 $\pm$ 0.7 & 13.6 $\pm$ 0.1 \\
    0.3 & 10 & 64.9 $\pm$ 0.2 & 18.2 $\pm$ 0.2 & 46.5 $\pm$ 0.2 & 56.3 $\pm$ 0.7 & 14.1 $\pm$ 0.1 \\
    0.5 & 10 & 64.9 $\pm$ 0.2 & 18.1 $\pm$ 0.2 & 46.5 $\pm$ 0.2 & 56.7 $\pm$ 0.7 & 14.3 $\pm$ 0.1 \\
    \midrule
    0.1 & 20 & 64.5 $\pm$ 0.2 & 18.0 $\pm$ 0.2 & 46.3 $\pm$ 0.2 & 54.8 $\pm$ 0.7 & 13.6 $\pm$ 0.1 \\
    0.3 & 20 & 65.1 $\pm$ 0.2 & 18.3 $\pm$ 0.2 & 46.7 $\pm$ 0.2 & 56.6 $\pm$ 0.7 & 14.3 $\pm$ 0.1 \\
    0.5 & 20 & 65.1 $\pm$ 0.2 & 18.2 $\pm$ 0.2 & 46.5 $\pm$ 0.2 & 57.1 $\pm$ 0.7 & 14.6 $\pm$ 0.1 \\
    \midrule
    0.1 & 30 & 64.6 $\pm$ 0.2 & 18.0 $\pm$ 0.2 & 46.3 $\pm$ 0.2 & 55.0 $\pm$ 0.7 & 13.7 $\pm$ 0.1 \\
    0.3 & 30 & 65.2 $\pm$ 0.2 & 18.3 $\pm$ 0.2 & 46.7 $\pm$ 0.2 & 56.9 $\pm$ 0.7 & 14.3 $\pm$ 0.1 \\
    0.5 & 30 & 64.9 $\pm$ 0.2 & 18.1 $\pm$ 0.2 & 46.7 $\pm$ 0.2 & 58.0 $\pm$ 0.7 & 14.7 $\pm$ 0.1 \\
    \bottomrule
    \end{tabular}
\end{sc}
\end{small}
\end{center}
\vskip -0.1in
\end{table}

\begin{table}[]
\caption{Comparison of different values for temperature of nucleus sampling on the MSRVTT validation set for $k=5$.}
\label{tab:temp_search_msrvtt}
\vskip 0.15in
\begin{center}
\begin{small}
\begin{sc}
    \begin{tabular}{c | c | c c c c c c c}
    \toprule
    Temperature & Samples & BLEU@1 & BLEU@4 & ROUGE-L & CIDEr-D & SPICE\\
    \midrule
    0.0 & n/a & 27.1 $\pm$ 0.1 & 4.9 $\pm$ 0.0 & 25.8 $\pm$ 0.1 & 36.7 $\pm$ 0.4 & 14.1 $\pm$ 0.1\\\midrule
    0.1 & 10 & 24.8 $\pm$ 0.1 & 4.4 $\pm$ 0.0 & 25.8 $\pm$ 0.1 & 37.4 $\pm$ 0.4 & 14.7 $\pm$ 0.1 \\
    0.3 & 10 & 24.9 $\pm$ 0.1 & 4.2 $\pm$ 0.0 & 25.6 $\pm$ 0.1 & 38.2 $\pm$ 0.4 & 14.8 $\pm$ 0.1 \\
    0.5 & 10 & 24.7 $\pm$ 0.1 & 4.1 $\pm$ 0.0 & 25.3 $\pm$ 0.1 & 37.9 $\pm$ 0.4 & 14.6 $\pm$ 0.1 \\
    \midrule
    0.1 & 20 & 24.7 $\pm$ 0.1 & 4.3 $\pm$ 0.0 & 25.7 $\pm$ 0.1 & 37.3 $\pm$ 0.4 & 14.7 $\pm$ 0.1 \\
    0.3 & 20 & 24.8 $\pm$ 0.1 & 4.2 $\pm$ 0.0 & 25.6 $\pm$ 0.1 & 38.0 $\pm$ 0.4 & 14.7 $\pm$ 0.1 \\
    0.5 & 20 & 24.6 $\pm$ 0.1 & 4.0 $\pm$ 0.0 & 25.3 $\pm$ 0.1 & 38.3 $\pm$ 0.4 & 14.6 $\pm$ 0.1 \\
    \midrule
    0.1 & 30 & 24.7 $\pm$ 0.1 & 4.3 $\pm$ 0.0 & 25.8 $\pm$ 0.1 & 37.3 $\pm$ 0.4 & 14.7 $\pm$ 0.1 \\
    0.3 & 30 & 24.7 $\pm$ 0.1 & 4.2 $\pm$ 0.0 & 25.6 $\pm$ 0.1 & 38.1 $\pm$ 0.4 & 14.7 $\pm$ 0.1 \\
    0.5 & 30 & 24.5 $\pm$ 0.1 & 4.0 $\pm$ 0.0 & 25.3 $\pm$ 0.1 & 38.1 $\pm$ 0.4 & 14.6 $\pm$ 0.1 \\
    \bottomrule
    \end{tabular}
\end{sc}
\end{small}
\end{center}
\vskip -0.1in
\end{table}

\begin{table}[]
\caption{Comparison of different values for temperature of nucleus sampling on the chest-xrays validation set for $k=5$.}
\label{tab:temp_search_xrays}
\vskip 0.15in
\begin{center}
\begin{small}
\begin{sc}
    \begin{tabular}{c | c | c c c c c c c}
    \toprule
    Temperature & Samples & BLEU@1 & BLEU@4 & ROUGE-L & CIDEr-D & SPICE\\
    \midrule
    0.0 & n/a & 30.7 $\pm$ 0.8 & 6.8 $\pm$ 0.4 & 30.9 $\pm$ 0.7 & 16.4 $\pm$ 1.4 & 10.3 $\pm$ 0.4 \\
    \midrule
    0.1 & 10 & 31.9 $\pm$ 0.8 & 7.1 $\pm$ 0.5 & 30.8 $\pm$ 0.7 & 15.8 $\pm$ 1.2 & 10.2 $\pm$ 0.4 \\
    0.3 & 10 & 33.5 $\pm$ 0.8 & 7.5 $\pm$ 0.5 & 31.2 $\pm$ 1.3 & 16.9 $\pm$ 1.3 & 10.3 $\pm$ 0.4 \\
    0.5 & 10 & 33.5 $\pm$ 0.8 & 7.4 $\pm$ 0.5 & 31.0 $\pm$ 0.7 & 17.3 $\pm$ 1.3 & 10.1 $\pm$ 0.4 \\
    0.7 & 10 & 33.1 $\pm$ 0.8 & 7.1 $\pm$ 0.5 & 31.1 $\pm$ 0.7 & 17.7 $\pm$ 1.3 & 10.1 $\pm$ 0.4 \\
    \midrule
    0.1 & 20 & 31.9 $\pm$ 0.8 & 7.2 $\pm$ 0.5 & 30.9 $\pm$ 0.7 & 15.9 $\pm$ 1.2 & 10.3 $\pm$ 0.4 \\
    0.3 & 20 & 33.1 $\pm$ 0.8 & 7.3 $\pm$ 0.5 & 31.3 $\pm$ 0.7 & 17.0 $\pm$ 1.3 & 10.2 $\pm$ 0.4 \\
    0.5 & 20 & 33.8 $\pm$ 0.8 & 7.4 $\pm$ 0.5 & 31.3 $\pm$ 0.7 & 18.2 $\pm$ 1.4 & 10.2 $\pm$ 0.4 \\
    0.7 & 20 & 32.4 $\pm$ 0.8 & 6.7 $\pm$ 0.5 & 30.6 $\pm$ 0.7 & 16.0 $\pm$ 1.3 & 10.2 $\pm$ 0.4 \\
    \midrule
    0.1 & 30 & 31.8 $\pm$ 0.8 & 7.2 $\pm$ 0.5 & 30.8 $\pm$ 0.7 & 15.8 $\pm$ 1.2 & 10.3 $\pm$ 0.4 \\
    0.3 & 30 & 33.0 $\pm$ 0.8 & 7.2 $\pm$ 0.5 & 31.2 $\pm$ 0.7 & 17.0 $\pm$ 1.3 & 10.1 $\pm$ 0.4 \\
    0.5 & 30 & 33.1 $\pm$ 0.8 & 7.3 $\pm$ 0.5 & 31.2 $\pm$ 0.7 & 18.2 $\pm$ 1.4 & 10.2 $\pm$ 0.4 \\
    0.7 & 30 & 32.6 $\pm$ 0.8 & 7.1 $\pm$ 0.5 & 30.8 $\pm$ 0.7 & 16.6 $\pm$ 1.2 & 10.1 $\pm$ 0.4 \\
    \bottomrule
    \end{tabular}
\end{sc}
\end{small}
\end{center}
\vskip -0.1in
\end{table}

The results for other decoding schemes are shown in \cref{tab:decoding_ablation}.
For greedy decoding we only generate one caption, hence no selection step is required after generation.
We use the same temperature as the best nucleus sampling setting for topk and regular sampling.
We find that nucleus sampling with $l=1$ performs close to greedy decoding, however when setting $l=10$ and using caption selection via our aligned CLIP, we observe a substantial improvement.

\begin{table}[]
\caption{Search over different decoding paradigms for captioning on the MS-COCO validation set. We report mean and standard error for all settings. Sampling-based decoding strategies use a temperature of $\tau=0.1$.}
\label{tab:decoding_ablation}
\vskip 0.15in
\begin{center}
\begin{small}
\begin{sc}
    \begin{tabular}{l c c c c c c c}
    \toprule
    Decoding & BLEU@1 & BLEU@4 & ROUGE-L & CIDEr-D & SPICE \\
    \midrule
    Sampling & 67.9 $\pm$ 0.2 & 21.0 $\pm$ 0.3 & 51.6 $\pm$ 0.2 & 80.7 $\pm$ 0.8 & 19.3 $\pm$ 0.1 \\
    Topk & 67.9 $\pm$ 0.2 & 20.8 $\pm$ 0.3 & 51.5 $\pm$ 0.2 & 80.9 $\pm$ 0.8 & 19.4 $\pm$ 0.1 \\
    Greedy & 77.4 $\pm$ 0.2 & 30.5 $\pm$ 0.4 & 57.7 $\pm$ 0.2 & 105.5 $\pm$ 1.0 & 20.8 $\pm$ 0.1 \\
    Nucleus, $l=1$ & 77.4 $\pm$ 0.2 & 30.4 $\pm$ 0.4 & 57.8 $\pm$ 0.2 & 105.5 $\pm$ 1.0 & 20.8 $\pm$ 0.1\\
    Nucleus & 77.7 $\pm$ 0.2 & 30.5 $\pm$ 0.4 & 58.0 $\pm$ 0.2 & 107.3 $\pm$ 1.0 & 21.2 $\pm$ 0.1 \\
    \bottomrule
    \end{tabular}
\end{sc}
\end{small}
\end{center}
\vskip -0.1in
\end{table}

\paragraph{Prompt ordering}

Usually we would provide the captions in the prompt from most-similar to least similar, i.e. the least similar prompt is the most recent in the context.
However, one may think the exact opposite ordering might lead to better captioning performance, since the LM might exhibit a form of recency bias.
This concerns our setting as well, since the values we found for $k$ are larger than one might expect, e.g., on MS-COCO we found $k=13$ to perform best.
Hence, we provide results for the worst-to-best ordering in \cref{tab:ordering_ablation}.
Indeed, we found that different ordering of captions in the prompt leads to different results.
Ordering from worst-to-best, i.e. most similar captions appear more recently, leads to an improvement on CIDEr-D score.
Therefore, by default, we provide the prompts in the order from worst-to-best in the prompt.

\begin{table}[]
\caption{Comparison of different orderings for exemplars in the prompt on the MS-COCO validation set. We report mean and standard error for all settings.}
\label{tab:ordering_ablation}
\vskip 0.15in
\begin{center}
\begin{small}
\begin{sc}
    \begin{tabular}{l c c c c c c c}
    \toprule
    Ordering & BLEU@1 & BLEU@4 & ROUGE-L & CIDEr-D & SPICE\\
    \midrule
    worst-to-best & 77.7 $\pm$ 0.2 & 30.5 $\pm$ 0.4 & 58.0 $\pm$ 0.2 & 107.3 $\pm$ 1.0 & 21.2 $\pm$ 0.1 \\
    best-to-worst & 77.4 $\pm$ 0.2 & 30.4 $\pm$ 0.4 & 57.7 $\pm$ 0.2 & 105.9 $\pm$ 1.0 & 21.0 $\pm$ 0.1 \\
    \bottomrule
    \end{tabular}
\end{sc}
\end{small}
\end{center}
\vskip -0.1in
\end{table}

\section{Motivation of Linear Alignment}
\label{app:motivation}

CLIP has been trained to align text with images in a joint embedding space. 
We want to use the CLIP encoders for retrieval by cosine similarity on an image-captioning task. 
However, there might be a disparity between the pretraining domain of CLIP and the downstream task. 
We aim to rectify this by a linear mapping. 
Our downstream task is retrieval of text embeddings $\Be_i$ by their corresponding image embeddings $\Bf_i$ using the cosine similarity. 
Therefore, our objective is
\begin{equation}
    \max_{\BW} \sum_i \operatorname{cossim}(\Be_i, \BW \Bf_i). \label{eqn:objective1}
\end{equation}
For objective \eqref{eqn:objective1} a closed-form solution is unknown. 
By constraining $\BW$ to be an orthogonal matrix, however, we obtain equivalence to the least-squares objective because
\begin{align}
    & \operatorname*{arg\,max}_{\BW^\top \BW = \BI} \sum_i \operatorname{cossim}(\Be_i, \BW \Bf_i) \label{eqn:proof1}\\
    =& \operatorname*{arg\,max}_{\BW^\top \BW = \BI} \sum_i \frac{\Be_i^\top \BW \Bf_i}{\|\Be_i\|_2 \|\BW \Bf_i\|_2} \label{eqn:proof2}\\
    =& \operatorname*{arg\,max}_{\BW^\top \BW = \BI} \sum_i \Be_i^\top \BW \Bf_i \label{eqn:proof3}\\
    =& \operatorname*{arg\,min}_{\BW^\top \BW = \BI} -\sum_i \Be_i^\top \BW \Bf_i \label{eqn:proof4}\\
    =& \operatorname*{arg\,min}_{\BW^\top \BW = \BI} \sum_i (\|\BW \Bf_i\|_2^2 + \|\Be_i\|_2^2 - 2 \Be_i^\top \BW \Bf_i) \label{eqn:proof5}\\
    =& \operatorname*{arg\,min}_{\BW^\top \BW = \BI} \sum_i (\Bf_i^\top \BW^\top \BW \Bf_i + \Be_i^\top \Be_i - 2 \Be_i^\top \BW \Bf_i) \label{eqn:proof6}\\
    =& \operatorname*{arg\,min}_{\BW^\top \BW = \BI} \sum_i (\BW \Bf_i - \Be_i)^\top (\BW \Bf_i - \Be_i) \label{eqn:proof7}\\
    =& \operatorname*{arg\,min}_{\BW^\top \BW = \BI} \sum_i \|\BW \Bf_i - \Be_i\|_2^2. \label{eqn:proof8}
\end{align}
\citet{artetxe_learning_2016} have pointed out this fact previously. 
Note that from \eqref{eqn:proof2} to \eqref{eqn:proof3} and from \eqref{eqn:proof4} to \eqref{eqn:proof5} the term $\|\BW \Bf_i\|_2$ can be dropped/added as it appears constant to the optimization objective because $\BW$ is orthogonal and, therefore, preserves the norm of $\Bf_i$. 
The solution to this optimization problem is known as orthogonal procrustes \citep{schonemann_generalized_1966} and can be written as 
\begin{equation}
    \BW = \BV \BU^\top,
\end{equation}
where $\BV$ and $\BU$ are the orthogonal matrices of the singular value decomposition of $\BF^\top \BE = \BU \BSi \BV^\top$ and $\BF = (\Bf_1, \dots, \Bf_n)^\top, \BE = (\Be_1, \dots, \Be_n)^\top$.

\end{document}